\documentclass{article} % For LaTeX2e
\usepackage{iclr2023_conference,times}

\usepackage{amsmath,amsfonts,bm}

\def\eqref#1{equation~\ref{#1}}
\def\1{\bm{1}}

\DeclareMathAlphabet{\mathsfit}{\encodingdefault}{\sfdefault}{m}{sl}
\SetMathAlphabet{\mathsfit}{bold}{\encodingdefault}{\sfdefault}{bx}{n}

\usepackage{hyperref}
\usepackage{url}
\usepackage{graphicx}
\usepackage{amsmath}
\usepackage{booktabs}
\usepackage{xcolor}
\usepackage{enumitem}

\newcommand{\algoname}[1]{Label-free CBM}
\newcommand{\yes}[1]{\textbf{\textcolor{blue}{Yes}}}
\newcommand{\rebuttal}[1]{\textcolor{black}{#1}}

\title{Label-free Concept Bottleneck Models}

\author{%
  Tuomas Oikarinen \\
  UCSD CSE \\
  \texttt{toikarinen@ucsd.edu} \\
  \And 
  Subhro Das \\
  MIT-IBM Watson AI Lab, IBM Research \\
  \texttt{subhro.das@ibm.com} \\
  \And 
  Lam M. Nguyen \\
  IBM Research\\
  \texttt{lamnguyen.mltd@ibm.com} \\
  \And
  Tsui-Wei Weng \\
  UCSD HDSI \\
  \texttt{lweng@ucsd.edu} 
  }

\iclrfinalcopy
\begin{document}

\maketitle

\begin{abstract}
Concept bottleneck models (CBM) are a popular way of creating more interpretable neural networks by having hidden layer neurons correspond to human-understandable concepts. However, existing CBMs and their variants have two crucial limitations:  first, they need to collect labeled data for each of the predefined concepts, which is time consuming and labor intensive; second, the accuracy of a CBM is often significantly lower than that of a standard neural network, especially on more complex datasets. This poor performance creates a barrier for adopting CBMs in practical real world applications. Motivated by these challenges, we propose \textit{Label-free} CBM which is a novel framework to transform any neural network into an interpretable CBM without labeled concept data, while retaining a high accuracy. Our Label-free CBM has many advantages, it is: \textit{scalable} - we present the first CBM scaled to ImageNet, \textit{efficient} - creating a CBM takes only a few hours even for very large datasets, and \textit{automated} - training it for a new dataset requires minimal human effort. Our code is available at \href{https://github.com/Trustworthy-ML-Lab/Label-free-CBM}{https://github.com/Trustworthy-ML-Lab/Label-free-CBM}. \textcolor{black}{Finally, in Appendix~\ref{sec:user_study} we conduct a large scale user evaluation of the interpretability of our method.}

\end{abstract}

\section{Introduction}
% \lily{\begin{itemize}
%     \item Paragraph 1: DNN is powerful but blackbox
%     \item Paragraph 2: Existing methods on interpreting DNN (post-hoc v.s. involved in training)
%     \item Paragraph 3: Limitation of existing CBM methods and bring out the motivation
%     \item Paragraph 4: Contribution
% \end{itemize}}

\begin{figure}[b!]
    \centering
    \includegraphics[width=0.58\linewidth]{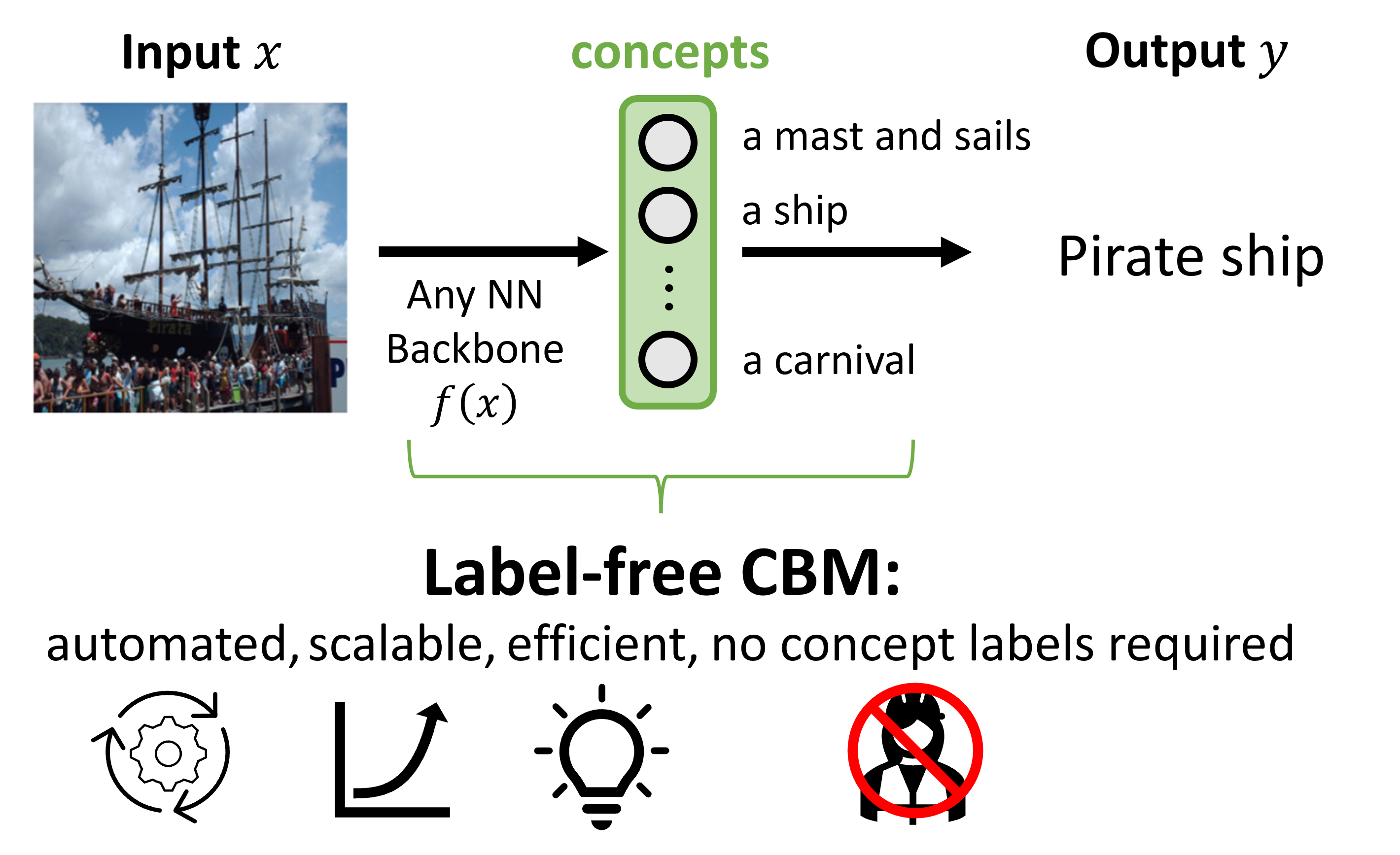}
    \caption{Our proposed \algoname{} has many desired features which existing CBMs lack, and it can transform any neural network backbone into an interpretable Concept Bottleneck Model.}
    \label{fig:our_CBM_feature}
\end{figure}

Deep neural networks (DNNs) have demonstrated unprecedented success in a wide range of machine learning tasks such as computer vision, natural language processing, and speech recognition. However, due to their complex and deep structures, they are often regarded as \textit{black-box} models that are difficult to understand and interpret. Interpretable models are important for many reasons such as creating calibrated trust in models, which means understanding when we should trust the models. Making deep learning models more interpretable is an active yet challenging research topic.

One approach to make deep learning more interpretable is through Concept Bottleneck Models (CBMs)~\citep{koh2020concept}. CBMs typically have a Concept Bottleneck Layer before the (last) fully connected layer of the neural network. The concept bottleneck layer is trained to have each neuron correspond to a single human understandable \textit{concept}. This makes the final decision a linear function of interpretable concepts, greatly increasing our understanding of the decision making. Importantly, CBMs have been shown to be useful in a variety of applications, including model debugging and human intervention on decisions. However, there are two crucial limitations of existing CBMs and their variants~\citep{koh2020concept,yuksekgonul2022post, zhou2018interpretable}: (i) labeled data is required for each of the predefined concepts, which is time consuming and expensive to collect; (ii) the accuracy of a CBM is often significantly lower than the accuracy of a standard neural network, especially on more complex datasets.  

To address the above two challenges, we propose a new framework named Label-free CBM, which is capable of transforming any neural network into an interpretable CBM \textit{without} labeled concept data while \textit{preserving} accuracy comparable to the original neural network by leveraging foundation models~\citep{foundation}. Our Label-free CBM has many advantages:
\begin{itemize}
    \item it is \textit{scalable} - to our best knowledge, it is the first CBM that scales to ImageNet
    \item it is \textit{efficient} - creating a CBM takes only a few hours even for very large datasets
    \item it is \textit{automated} - training it for a new dataset requires minimal human effort
\end{itemize}

\section{Related work}

\textbf{Post-hoc explanations}~\citep{samek2021explaining}: The approach of post-hoc explanations includes some classic methods such as LIME~\citep{ribeiro2016should} and SHAP~\citep{lundberg2017unified}, which try to explain individual model decisions by identifying which parts of the input data (e.g. pixels) are the most important for a given decision. However, these methods are based on local approximations of the DNN model and as such are not always accurate. Further, the explanations at the granularity of input pixels may not always be helpful and could require substantial subjective analysis from human. In contrast, our explanations in Section \ref{sec:experiment} are not approximated and explain predictions in terms of human-understandable concepts.

\textbf{More interpretable final layer}: \citep{wong2021leveraging} proposes making the FC layer sparse, and develop an efficient algorithm for doing so. They show that sparse models are more interpretable in many ways, but it still suffers from the fact the previous layer features are not interpretable. NBDT \citep{wan2020neural} propose replacing the final layer with a neural backed decision tree for another form of more interpretable decisions. Other approaches to make NNs more interpretable include Concept Whitening \citep{chen2020concept} and Concept Embedding Models \citep{zarlenga2022concept}.

\textbf{CBM}: Most related to our approach are Concept Bottleneck Models~\citep{koh2020concept, losch2019interpretability} which create a layer before the last fully connected layer where each neuron corresponds to a human interpretable concept. CBMs have been shown to be beneficial by allowing for human test-time intervention for improved accuracy, as well as being easier to debug.
To reduce the training cost of a CBM, a recent work \citep{yuksekgonul2022post} proposed Post-Hoc CBM that only needs to train the last FC layer along with an optional residual fitting layer, avoiding the need to train the backbone from scratch. This is done by leveraging Concept Activation Vectors (CAV)~\citep{kim2018interpretability} or the multi-modal CLIP model~\citep{clip}. However, the post-hoc CBM does not fully address the problems of the original CBM as using TCAV still requires collecting annotated concept data and their use of CLIP model can only be applied to if the NN backbone is the CLIP image encoder. Additionally, the performance of post-hoc CBMs without uninterpretable residual fitting layers is often significantly lower than the standard DNNs. Similarly, an earlier work Interpretable Basis Decomposition \citep{zhou2018interpretable} proposes learning a concept bottleneck layer based on labeled concept data for explanable decisions, even though they do not call themselves a CBM. Comparison between the features our method and existing approaches is shown in Table \ref{tab:pros_cons}. 

% \usepackage{booktabs}
\begin{table}[t]
\centering
\scalebox{0.8}{
\begin{tabular}{@{}l||cc|cc|cc@{}}
\toprule
 & \multicolumn{2}{c}{(I) Flexibility} & \multicolumn{2}{c}{(II) Interpretability} & \multicolumn{2}{c}{(III) Performance} \\ \midrule
Method: & \begin{tabular}[c]{@{}c@{}}Without labeled\\ concept data\end{tabular} & \begin{tabular}[c]{@{}c@{}}Any network\\ architecture\end{tabular} & \begin{tabular}[c]{@{}c@{}}Sparse \\final layer\end{tabular} & \begin{tabular}[c]{@{}c@{}} All features\\  interpretable\end{tabular} & \begin{tabular}[c]{@{}c@{}}Preserves\\  accuracy\end{tabular} & \begin{tabular}[c]{@{}c@{}}Extends to \\ ImageNet scale\end{tabular} \\ \midrule
CBM & No & \textbf{Yes} & No & \textbf{Yes} & No & No \\
IBD & No & \textbf{Yes} & No & No & \textbf{Yes} & No \\
P-CBM & No & \textbf{Yes} & Yes & \textbf{Yes} & No & No \\
P-CBM (CLIP) & \textbf{Yes} & No & \textbf{Yes} & \textbf{Yes} & No & Maybe \\
P-CBM-h & No & \textbf{Yes} & \textbf{Yes} & No & \textbf{Yes} & No \\
P-CBM-h (CLIP) & \textbf{Yes} & No & \textbf{Yes} & No & \textbf{Yes} & Maybe \\ \midrule
\begin{tabular}[c]{@{}l@{}} \algoname{} \\ \textbf{(This work)}\end{tabular} & \textbf{Yes} & \textbf{Yes} & \textbf{Yes} & \textbf{Yes} & \textbf{Yes} & \textbf{Yes} \\ \bottomrule
\end{tabular}
}
\caption{Comparison of our method against existing methods for creating Concept Bottleneck models, CBM \citep{koh2020concept}, IBD \citep{zhou2018interpretable} and 4 versions of P-CBM \citep{yuksekgonul2022post}, where `-h` indicates the hybrid model that uses uninterpretable residual term, `(CLIP)` means models using CLIP concepts. We used maybe to indicate models that could in theory extend to ImageNet but have not been tested.}
\label{tab:pros_cons}
\end{table}

\textbf{Model editing/debugging}: Our approach is related to a range of works proposing ways to edit networks, such as \citep{bau2020rewriting, wang2022rewriting} for generative vision models, \citep{bau2020rewriting} for classifiers, or \citep{meng2022locating, fastedit} for language models. In addition \citep{abid2021meaningfully} propose a way to debug model mistakes using TCAV activation vectors.

\textbf{CLIP-Dissect}~\citep{oikarinen2022clip}: CLIP-Dissect is a recent method for understanding the roles of hidden layer neurons by leveraging the CLIP multimodal model \citep{clip}. It can provide a score of how close any neuron is to representing any given concept without the need of concept annotation data, which makes it useful as an optimization target for learning our interpretable projection in Step 3.

\begin{figure}[b]
    \centering
    \includegraphics[width=0.99\linewidth]{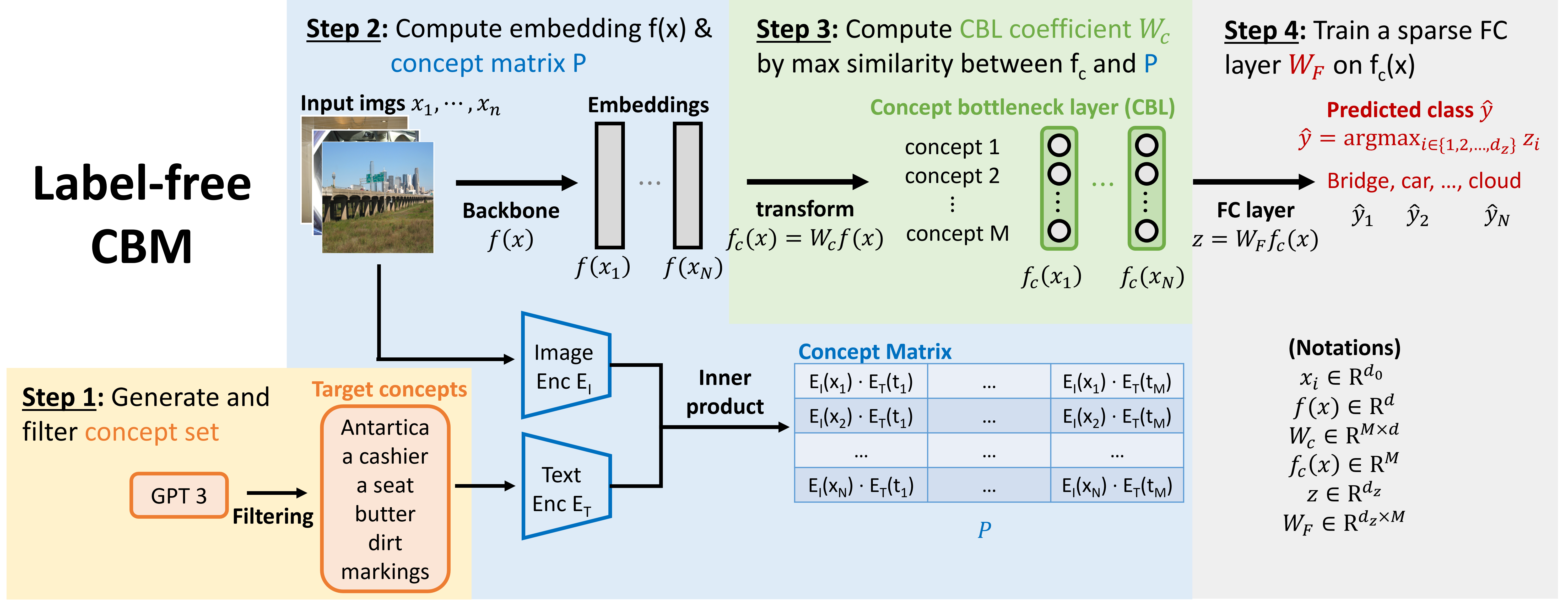}
    \caption{Overview of our pipeline for creating label-free CBM.}
    \label{fig:schematic}
\end{figure}

\section{Label-free CBM: A new framework to build CBM}
In this section, we propose \algoname{}, a novel framework that builds a concept bottleneck model (CBM) in an \textit{automated}, \textit{scalable} and \textit{efficient} fashion and addresses the core limitations of existing CBMs. Given a neural network backbone, \algoname{} transforms the backbone into an interpretable CBM without the need of concept labels with the following 4 steps which are illustrated in Fig \ref{fig:schematic} -- \textbf{Step~1:} Create the initial concept set and filter undesired concepts; \textbf{Step~2:} Compute embeddings from the backbone and the concept matrix on the training dataset; \textbf{Step~3:} Learn projection weights $W_c$ to create a Concept Bottleneck Layer (CBL); \textbf{Step~4:} Learn the weights $W_F$ of the sparse final layer to make predictions.

Note that the backbone model can either be a model trained on the target task, or a general model trained on a different task.  
The details of \algoname{} for Step~1 are provided in Sec~3.1, Step~2 and Step~3 in Sec~3.2, and finally Step~4 in Sec~3.3. 

\subsection{Step 1: Concept set creation and filtering}

\label{sec:concept_set}

In this step, we will describe how to create a concept set to serve as the basis of human-interpretable concepts in the Concept Bottleneck Layer. This step consists of two sub-steps: \textbf{A. Initial concept set creation} and \textbf{B. Concept set filtering}.

\textbf{A. Initial concept set creation}: A concept set refers to the set of concepts represented in the Concept Bottleneck Layer. In the original CBM paper \citep{koh2020concept}, this is decided by domain experts as the set of concepts that are important for the given task. However, since our objective is to automate the entire process of generating CBMs, we don't want to rely on human experts. Instead, we propose generating the concept set via GPT-3 \citep{gpt3} using the OpenAI API. Somewhat surprisingly, GPT-3 has a good amount of domain knowledge of which concepts are important for detecting each class when prompted in the right way. Specifically, we ask GPT-3 the following:
\begin{itemize}
    \item List the most important features for recognizing something as a \{class\}:
    \item List the things most commonly seen around a \{class\}:
    \item Give superclasses for the word \{class\}:
\end{itemize}

Note that the \{class\} in here refers to the class name in the machine learning task. For GPT-3 to perform well on the above prompt, we provide two examples of the desired outputs for few-shot adaptation. Note that those two examples can be shared across all datasets, so no additional user input is needed for generating concept set for a new dataset. Full prompts and example outputs are illustrated in the Appendix Figures \ref{fig:gpt_1} and \ref{fig:gpt_2}. To reduce variance we run each prompt twice and combine the results. Combining the concepts received from different classes and prompts gives us a large, somewhat noisy set of initial concepts, which we further improve by filtering. We found using GPT-3 to generate initial concepts to perform better than using the knowledge-graph ConceptNet \citep{conceptnet} which was used in Post Hoc-CBM \citep{yuksekgonul2022post}. Reasons for this and a comparison between the two are presented in Appendix \ref{sec:conceptnet_comparison}.

\textbf{B. Concept set filtering}: Next we employ several filters to improve the quality and reduce the size of our concept set, as stated below:
\begin{enumerate}
    \item \textit{Concept length}: We delete any concept longer than 30 characters in length, to keep concepts simple and avoid unnecessary complication.
    \item \textit{Remove concepts too similar to classes}: We don't want our CBM to contain the output classes themselves as that would defeat the purpose of an explanation. To avoid this we remove all concepts that are too similar to the names of target classes. We measure this with cosine similarity in a text embedding space. In particular we use an ensemble of similarities in the CLIP ViT-B/16 text encoder as well as the \href{https://www.sbert.net/docs/pretrained_models.html}{all-mpnet-base-v2} sentence encoder space, so our measure can be seen as a combination of visual and textual similarity. For all datasets, we deleted concepts with similarity $>0.85$ to any target class.
    \item \textit{Remove concepts too similar to each other}: We also don't want duplicate or synonymous concepts in the bottleneck layer. We use the same embedding space as above, and remove any concept that has another concept with $>0.9$ cosine similarity already in concept set.
    \item \textit{Remove concepts not present in training data}: To make sure our concept layer accurately presents its target concepts, we remove any concepts that don't activate CLIP highly. This cut-off is dataset specific, and we delete all concepts with average top-5 activation below the cut-off.
    \item \textit{Remove concepts we can't project accurately}: Remove neurons that are not interpretable from the CBL. This step is actually performed after step 3 and is described in section \ref{sec:method_CBL}.
\end{enumerate}

We discuss the reasoning behind the filters in detail in Appendix \ref{sec:filter_example} and perform an abblation study on their effects in Appendix \ref{sec:filter_ablation}.

\subsection{Step 2 and 3: Learning the Concept Bottleneck Layer (CBL)}
\label{sec:method_CBL}
Once the concept set is obtained, the next step is to learn a projection from the backbone model's feature space into a space where axis directions correspond to interpretable concepts. Here, we present a way of learning the projection weights $W_c$ without any labeled concept data by utilizing CLIP-Dissect \citep{oikarinen2022clip}. To start with, we need a set of target concepts that the bottleneck layer is expected to represent as $\mathcal{C} = \{ t_1, ..., t_M\}$, as well as a training dataset (e.g. images) $\mathcal{D}=\{x_1, ..., x_N\}$ of the original task. Next we calculate and save the CLIP concept activation matrix $P$ where $P_{i,j} = E_I(x_i) \cdot E_T(t_j)$ and $E_I$ and $E_T$ are the CLIP image and text encoders respectively. $W_c$ is initialized as a random $M \times d_0$ matrix where $d_0$ is the dimensionality of backbone features $f(x)$. The initial set $\mathcal{C}$ is created in Step 1 and the training set $\mathcal{D}$ is provided by the downstream task. We define $f_c(x) = W_c f(x)$, where $f_c(x_i) \in \mathbb{R}^M$. We use $k$ to denote a neuron of interest in the projection layer, and its activation pattern is denoted as $q_k$ where $q_k = [f_{c, k}(x_1), \ldots, f_{c, k}(x_N)]^\top$, with $q_k \in \mathbb{R}^N$ and $f_{c,k}(x) = [f_c(x)]_k$. 

To make the neurons in the CBL interpretable, we need to enforce the projected neurons to activate in correlation with the target concept, which we do by optimizing $W_c$ to maximize the CLIP-Dissect similarity between the neuron's activation pattern and the target concept. To optimize this similarity, we have designed a new fully differentiable similarity function $\textrm{sim}(t_i, q_i)$ that can be applied to CLIP-Dissect, called \textit{cos cubed}, that still achieves very good performance in explaining the neuron functionality as shown in Appendix \ref{sec:cos_cubed_performance}. Our optimization goal is to minimize the objective $L$ over $W_c$ as defined in Equation~(\ref{eq:loss}): 
\begin{equation}
\label{eq:loss}
    L(W_c) = \sum_{i=1}^{M}-\textrm{sim}(t_i, q_i) := \sum_{i=1}^{M}-\frac{\bar{q_i}^3 \cdot \bar{P_{:, i}}^3}{||\bar{q_i}^3||_2 ||\bar{P_{:, i}}^3||_2}.
\end{equation}
Here $\bar{q}$ indicates vector $q$ normalized to have mean 0 and standard deviation 1, and the \textit{cos cubed} similarity $\textrm{sim}(t_i, q_i)$ is simply the cosine similarity between two activation vectors after both have been normalized and raised to third power element-wise. The third power is necessary to make the similarity more sensitive to highly activating inputs. As this is still a cosine similarity, it takes values between $[-1, 1]$.
We optimize $L(W_c)$ using the Adam optimizer on training data $\mathcal{D}$, with early stopping when similarity on validation data starts to decrease. Finally to make sure our concepts are truthful, we drop all concepts $j$ with $\textrm{sim}(t_j, q_j)<0.45$ on validation data after training $W_c$. This is the 5th concept set filter from Sec 3.1. This cutoff was selected manually as a good indicator of a neuron being interpretable. During this filtering, the number of concept is reduced: $M \leftarrow M - \Delta$, where $\Delta$ is non-negative integer representing the number of concepts being removed in this step. Note that matrix $W_c$ has also to be updated accordingly by removing the rows that corresponds to the removed concepts, and with our notation, $ W_c \in \mathbb{R}^{M \times d_0}$. To simplify Figure~\ref{fig:schematic}, we omit plotting this concept removal step.

\subsection{Step 4: Learning the sparse final layer}

Now that the Concept Bottleneck Layer is learned, the next task is to learn the final predictor with the fully connected layer $W_F$ $\in \mathbb{R}^{d_z \times M}$ where $d_z$ is the number of output classes. The goal is to keep $W_F$ sparse, since sparse layers have been demonstrated to be more interpretable~\citep{wong2021leveraging}. Given that both the backbone $f(x)$ and learned concept projection $W_c$ are fixed, this is a problem of learning a sparse linear model, which can be solved efficiently with the elastic net objective:
\begin{equation}
\label{eq:sparse}
    \min_{W_F, b_{F}} \sum_{i=1}^N L_{ce} (W_F f_c(x_i)  + b_{F}, y_i) + \lambda R_{\alpha}(W_F)
\end{equation}
where $R_{\alpha}(W_F) = (1-\alpha)\frac{1}{2}||W_F||_F + \alpha ||W_F||_{1,1}$, $\|\cdot\|_F$ denotes the Frobenius norm, $\|\cdot\|_{1,1}$ denotes element wise matrix norm, $b_{F}$ denotes the bias of the FC layer, $L_{ce}$ is the standard cross-entropy loss and $y_i$ is the ground-truth label of data $x_i$.
We optimize Equation~(\ref{eq:sparse}) using the GLM-SAGA solver created by \citep{wong2021leveraging}. For the sparse models, we used $\alpha=0.99$ and $\lambda$ was chosen such that each model has $25$ to $35$ nonzero weights per output class. This level was found to still result in interpretable decisions while retaining good accuracy. Depending on the number of features/concepts in the previous layer this corresponds to $0.7$-$15\%$ of the weights of the model being nonzero.

\section{Experiment Results}
\label{sec:experiment}
We present three main results on evaluating the accuracy and interpretability of the \algoname{} in this section. Due to page limit, additional experiments and discussions are in Appendix~\ref{sec:app:limitation}-\ref{sec:random_input_explanations}. An overview of additional experiments is provided in Appendix~\ref{sec:app:overview}. 

\textbf{Datasets.}
To evaluate our approach, we train Label-free CBMs on 5 datasets. These are CIFAR-10, CIFAR-100 \citep{krizhevsky2009learning}, CUB \citep{WahCUB_200_2011}, Places365 \citep{zhou2017places} and ImageNet \citep{imagenet}. This is a diverse set of tasks, where CIFAR-10/100 and ImageNet are general image classification datasets, CUB is a fine-grained bird-species classification dataset and Places365 is focused on scene recognition. Their sizes vary greatly, with CUB having 5900 training samples, CIFAR datasets 50,000 each and ImageNet and Places365 have 1-2 million training images. CUB contains annotations with 312 concepts for each datapoint, such as \textit{has wing color:blue} or \textit{has head pattern:spotted}. We used neither the concept names or the concept annotations to train our LF-CBM models to showcase our ability to perform without labels, yet our method discovered similar concepts and is competitive with methods that utilize the available concept information.

\textbf{Setup.} For CIFAR and CUB we use the same backbone models as \citep{yuksekgonul2022post} for fair comparison, so we use CLIP(RN50) image encoder as the backbone for CIFAR, and ResNet-18 trained on CUB from \href{https://github.com/osmr/imgclsmob}{imgclsmob} for CUB. For both ImageNet and Places365 we use ResNet-50 trained on ImageNet as the backbone network. The number of concepts each model uses is roughly proportional to the number of output classes for that task, as each class adds more initial concepts. The number of concepts for our models are as follows: 128 for CIFAR-10, 824 for CIFAR-100, 211 for CUB, 2202 for Places-365 and 4505 for ImageNet. CUB has a smaller number of concepts because we only used the \textit{important features} prompt. All models are trained on a single Nvidia Tesla P100 GPU, and the full training run takes anywhere from few minutes to 20 hours depending on the dataset size. The majority of runtime is taken by step 2, where we save activations for both the backbone and CLIP over the entire training dataset. Fortunately, these results only need to be calculated once and can be reused for training new CBMs on the same dataset. Once the activations have been saved, learning the model takes less than 4 hours on all datasets.

\textbf{Result (I): Accuracy.}
Table \ref{tab:accuracy} shows the performance of \algoname{} on all 5 datasets. We can see that our method can create a CBM with a sparse final layer on  with little loss in accuracy on all datasets, including a model with 72\% top-1 accuracy on the ImageNet. Our \algoname{} has significantly higher accuracy than Post-hoc CBM on the datasets we evaluated, but some rows for P-CBM are N/A as they do not provide results those results and it is unclear how to scale P-CBM to larger datasets. For Table \ref{tab:accuracy} we excluded methods with non-interpretable components (i.e. P-CBM-h or IBD) as we want to focus on fully interpretable CBM models. Note that P-CBM uses the expert provided concept set for CUB200, yet we can outperform it using our fully GPT-3 derived set of features. The standard sparse models were finetuned by us by learning a sparse final layer directly after feature layer $f(x)$ as described in \citep{wong2021leveraging} and also have 25-35 nonzero weights per class. The accuracies of sparse standard models are comparable to our CBM, indicating the CBM does not reduce accuracy.

\begin{table}[h!]
\centering
\scalebox{0.82}{
\begin{tabular}{@{}lcrrrcc@{}}
\toprule
 &  & \multicolumn{1}{l}{} & \multicolumn{1}{l}{} & \multicolumn{1}{c}{Dataset} &  &  \\ \cmidrule{3-7}
Model & Sparse final layer & \multicolumn{1}{l}{CIFAR10} & \multicolumn{1}{l}{CIFAR100} & \multicolumn{1}{l}{CUB200} & Places365 & ImageNet \\ \midrule
Standard & No & 88.80\%* & 70.10\%* & 76.70\% & \multicolumn{1}{r}{48.56\%} & \multicolumn{1}{r}{76.13\%} \\ \midrule
Standard (sparse) & Yes & 82.96\% & 58.34\% & \textbf{75.96\%} & \multicolumn{1}{r}{38.46\%} & \multicolumn{1}{r}{\textbf{74.35\%}} \\
P-CBM & Yes & 70.50\%* & 43.20\%* & 59.60\%* & N/A & N/A \\
P-CBM (CLIP) & Yes & 84.50\%* & 56.00\%* & \multicolumn{1}{c}{N/A} & N/A & N/A \\ \midrule
\begin{tabular}[c]{@{}l@{}}\algoname{}\\ \textbf{(Ours)}\end{tabular} & Yes & 
\begin{tabular}[c]{@{}r@{}}\textbf{86.40\%}\\ $\pm$ 0.06\% \end{tabular} & \begin{tabular}[c]{@{}r@{}}\textbf{65.13\%}\\ $\pm$ 0.12\% \end{tabular} & 
\begin{tabular}[c]{@{}r@{}}74.31\%\\ $\pm$ 0.29\% \end{tabular} &
\begin{tabular}[c]{@{}r@{}}\textbf{43.68\%}\\ $\pm$ 0.10\% \end{tabular} & \begin{tabular}[c]{@{}r@{}}71.95\%\\ $\pm$ 0.05\% \end{tabular} \\ \bottomrule
\end{tabular}
}
\caption{Accuracy comparison, best performing sparse model bolded. We can see our method outperforms Posthoc-CBM and performs similarly to a sparse standard model. The results for our method are mean and standard deviation over three training runs. *Indicates reported accuracy.}
\label{tab:accuracy}
\end{table}

\begin{figure}[t!]
    \centering
    \includegraphics[width=0.95\linewidth]{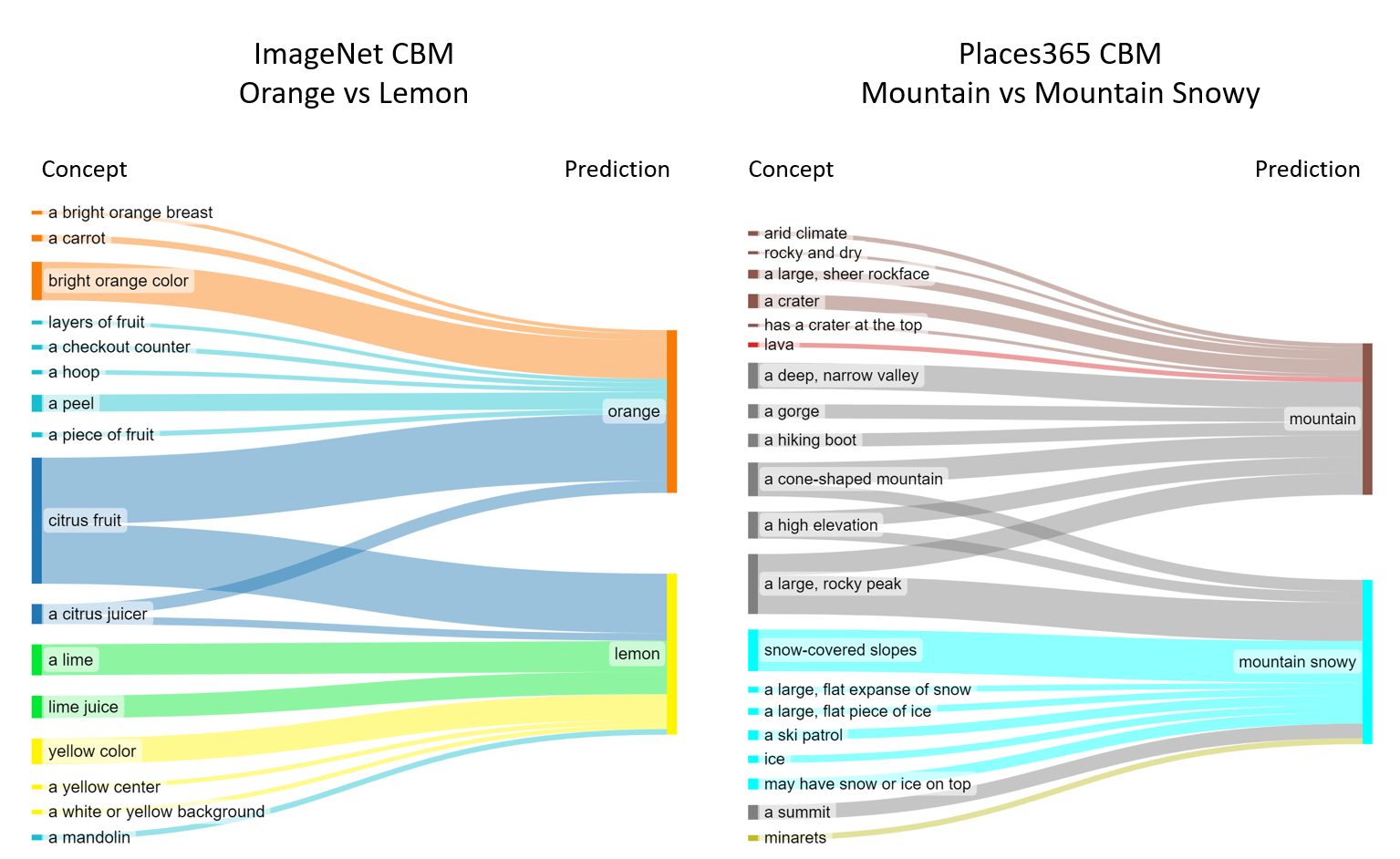}
    \caption{Visualization of the final layer weights of our \algoname{}. Showcasing how our models differentiate between two similar classes. }
    \label{fig:weights}
\end{figure}

\textbf{Result (II): Explainable decision rules.}
Perhaps the biggest benefit of Concept Bottleneck Models is that their decisions can be explained as a simple linear combination of understandable features. To showcase this, in Figure \ref{fig:weights} we visualize the final layer weights for the classes ``Orange" and ``Lemon" on ImageNet, and classes ``Mountain" and ``Mountain Snowy" on Places365. This visualization is a Sankey diagram of the final layer weights coming into two output classes, with the weight between a concept and output class displayed as the width of the line connecting them. We have only included weights with absolute value greater than 0.05 (for comparison the largest weights are usually 0.5-1). Negative weights are denoted as NOT {concept}. The learned decision rules largely align with our intuitions. For ImageNet, the concept ``citrus" fruit is highly connected to both ``Orange" and ``Lemon", while orange colors activate the ``Orange" class and yellow colors and limes activate the ``Lemon" class. On Places, concepts like ``a high elevation" activate both classes, while ``Mountain Snowy" is activated by snow and ice related concepts and ``Mountain" is activated by concepts related to volcanoes like ``lava" and ``crater". Visualizations like this allows us to gain a \textit{global} understanding of how our model behaves. Additional visualizations are shown in Appendix \ref{sec:app:extra_weights}

\textbf{Result (III): Explainable individual decisions.}
% \label{sec:explanation}
%
In addition to global understanding, CBMs allow us to understand the reasoning behind individual decisions. Since our decisions are linear functions of interpretable features, they can be explained in a simple and accurate way. After learning the Concept Bottleneck Layer, we normalize the activations of each concept to have mean 0 and standard deviation 1 on the training data. Given normalized features, the contribution of feature $j$ to output $i$ on input $x_k$ can be naturally calculated as $\textit{Contrib}(x_k,i,j) = W_{F[ij]} f_c(x_k)_j$. Since our $W_F$ is sparse, most contributions will be 0 and the important contributions can be easily visualized with a bar plot. Example visualizations for explaining the models predicted class are shown in Figure \ref{fig:contributions}, and more visualizations are available in Appendix \ref{sec:random_input_explanations}. These visualizations show the concepts with highest absolute value contribution. Concepts with negative activation can still be important contributors, and they are shown as ``NOT {concept}" in our visualizations. We can see for example the CUB model correctly identifies ``a red head" as the most important feature in recognizing this image as ``Red headed Woodpecker", while the fact that there is no ``a rosy breast" present also increases models confidence.

\begin{figure}[h!]
    \centering
    \includegraphics[width=0.85\linewidth]{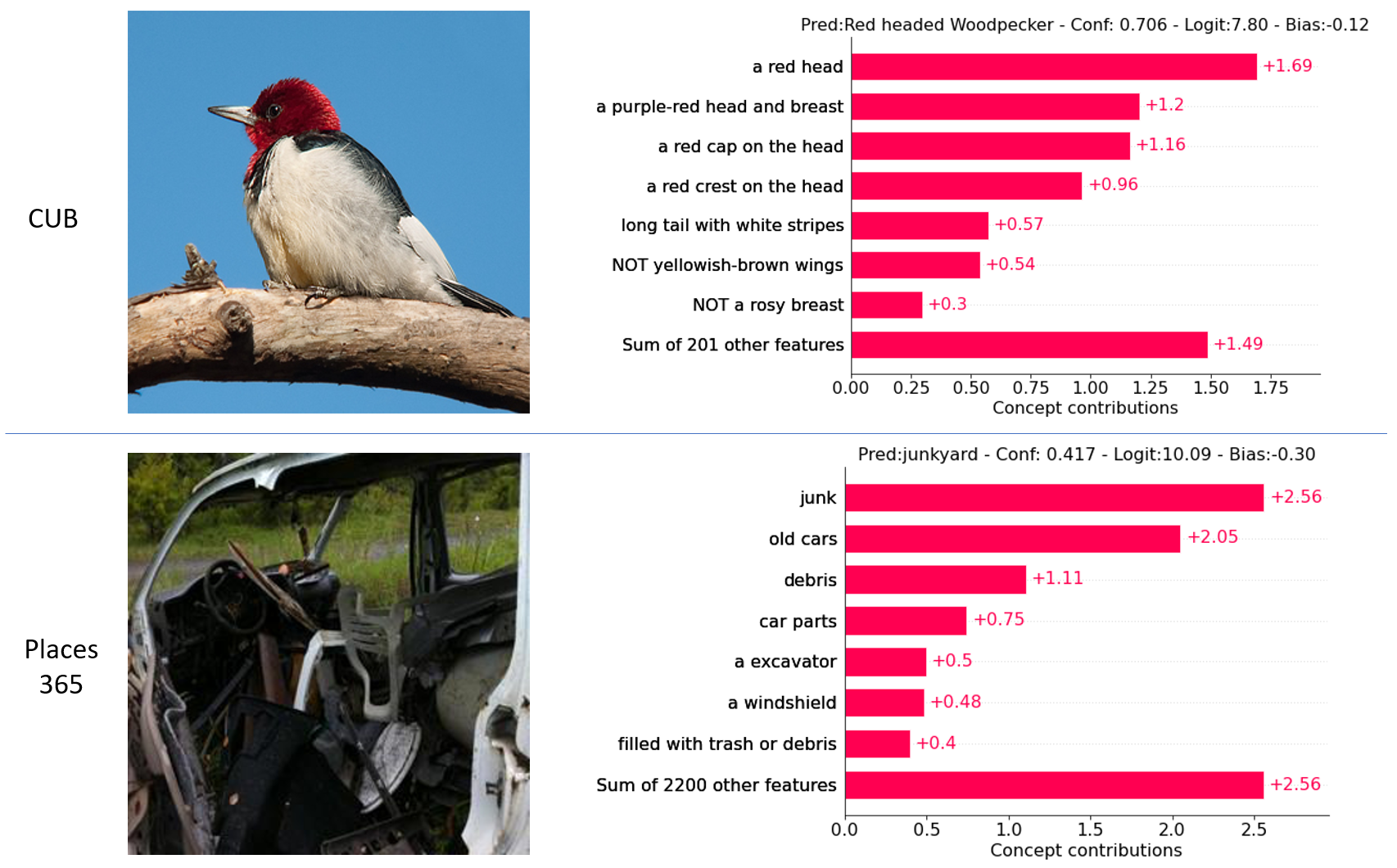}
    \caption{Visualization of two correct decisions made by our \algoname{}.}
    \label{fig:contributions}
\end{figure}

\vspace{-2mm}
\section{Case study: Manually improving an ImageNet model}
\vspace{-2mm}
In this section we take a deep dive into our ImageNet \algoname{} -- we analyze the errors it makes and design ways to debug/improve our network based on these findings. In particular, we show how we can inspect a handful of individual incorrect decisions and manually change our model weights to not only fix those predictions but also multiple other predictions, improving the networks overall accuracy. To our knowledge, this is the first example of manually editing a large well-trained neural network in a way that improves test accuracy on the same distribution.
\vspace{-2mm}
\subsection{Types of model errors}
\vspace{-2mm}
During the course of our experiment, we were able to identify 4 different types of mistakes our model makes, with different strategies for fixing them:

\textbf{Type 1: Incorrect/ambiguous label}: Despite looking like errors in terms of accuracy, these are simply a consequence of having noisy labels and are unavoidable. We will not attempt to fix this type of mistakes.

\textbf{Type 2: No sufficient concept in CBL}: Despite having a large set of concepts, our CBL does not include all concepts important for detecting certain classes. For example, detecting the difference between two species of snakes may require recognizing very fine-grained patterns that can't be easily explained in words. While these could often be fixed by adding new concepts to the concept set, we do not focus on it as it requires retraining the projection $W_{c}$ and the final layer.

\begin{figure}
    \centering
    \includegraphics[width=0.8\linewidth]{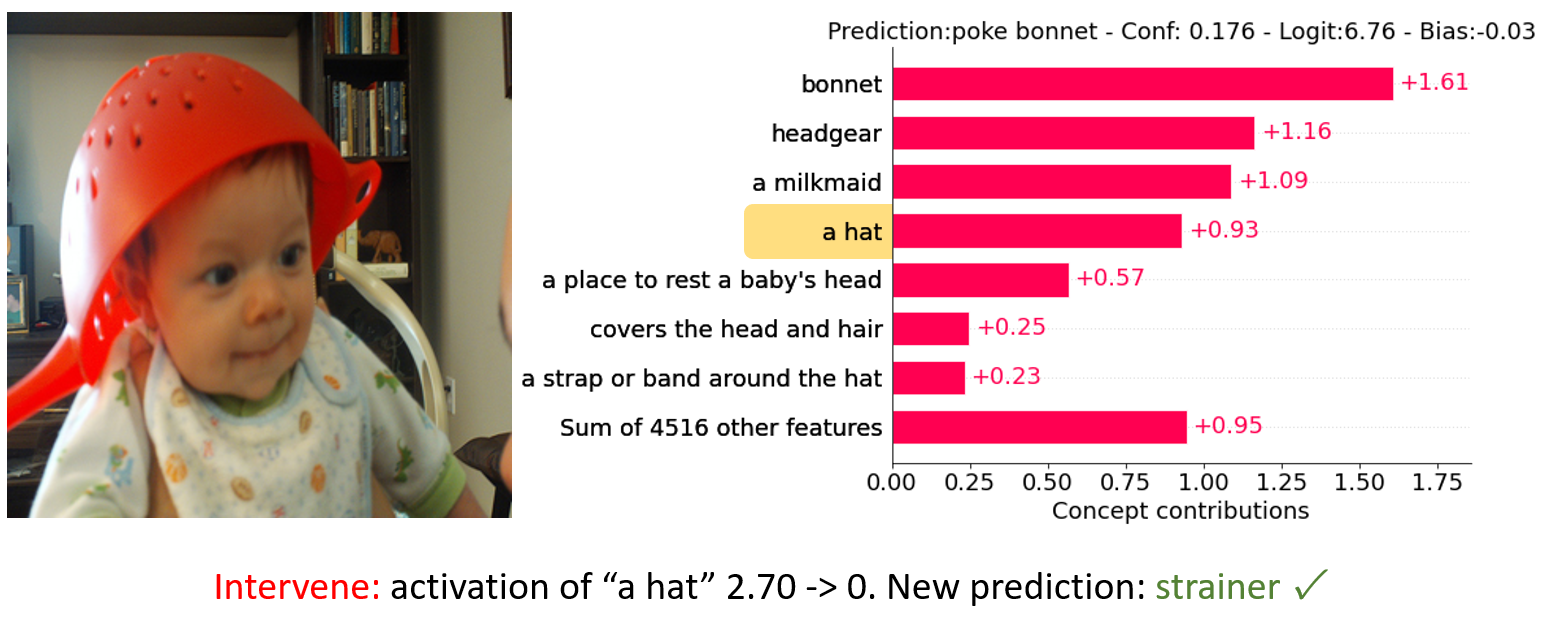}
    \caption{Here we fix an incorrect model prediction on ImageNet by simply zeroing out the incorrect activation for concept "a hat". Model originally predicted "poke bonnet" which is a type of hat.}
    \label{fig:intervention}
\end{figure}

\textbf{Type 3: Incorrect concept activations}: Some of the time, the activations in our CBL are incorrect and do not match the image, causing the prediction to become incorrect. Since the network up to the Concept Bottleneck Layer is mostly a black box, we can't easily improve the predictions, but predictions can be improved with test time interventions as shown in Figure \ref{fig:intervention}.

\textbf{Type 4: Incorrect final layer weight $W_F$}: Sometimes even if all the concept activations are correct, the models final layer weights still cause it to  make an erroneous prediction. As the final layer is fully interpretable, fixing these errors is the main focus of our study described in Sec \ref{sec:editing_weights}.

Examples of Type 1 and Type 2 errors are shown in Figure \ref{fig:error_types} in the Appendix.

\vspace{-2mm}
\subsection{Editing final layer weights}
\label{sec:editing_weights}
\vspace{-2mm}
In this section we describe our procedure for manually editing final layer weights of our model. This is done with the following procedure:
\begin{enumerate}[leftmargin=*]
    \item \textbf{Find an input where the model makes a Type 4 error}: This is done by visualizing incorrect model predictions and their explanations, and identifying Type 4 errors, i.e. errors where the highly activating concepts look correct and the label is correct and unambiguous.
    \item \textbf{Identify a concept to edit.}: Perhaps the most tricky step, to change a prediction we need to select a concept that is highly activated in this image, important for the ground truth class and not important for the incorrect (using our domain knowledge), and will have a small impact on other predictions. For example to flip a prediction from "cricket insect" to "grasshopper" we chose the concept "a green color".
    \item \textbf{Change weights by sufficient magnitude}: Next we choose magnitude of weight change $\Delta w$, and edit the weights in the following way: $W_{F[gt, concept]} \leftarrow W_{F[gt, concept]} + \Delta w$ and $W_{F[pred, concept]} \leftarrow W_{F[pred, concept]} - \Delta w$. Typically we want $\Delta w$ to be slightly more than required to flip the prediction on our example, but understanding its effect on other predictions is hard and should be checked with a validation dataset. 
\end{enumerate}

\begin{figure}
    \centering
    \includegraphics[width=0.8\linewidth]{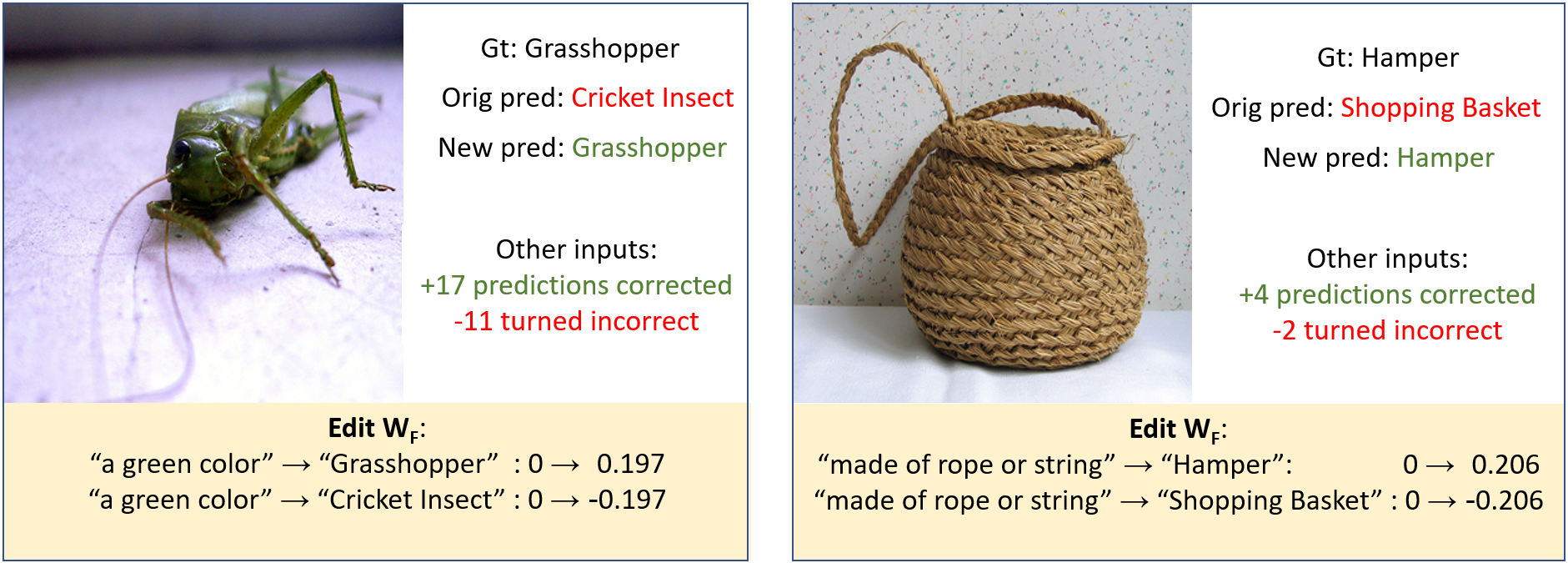}
    \caption{Examples of two edits we performed on our model predictions. Both edits fix the error we found, and fix more other predictions than they cause mistakes on other unseen inputs.}
    \label{fig:model_editsl}
\end{figure}

Fig \ref{fig:model_editsl} shows two examples of model edits we performed, and their effects on model predictions on the validation set. In total we identified 5 such beneficial edits during our quick exploration, the other 3 are shown in Fig \ref{fig:appendix_edits}. After applying all of them, the total validation accuracy of our model goes from 71.98\% to 72.02\%, correcting 38 predictions while turning 17 incorrect. While this might seem like a small difference, it is worth noting we only change 10/4.5 million total weights in the final layer, of which 34,550 are non-zero. Our edits only affect 10 classes which have a total of 500 validation examples, so in effect we increase accuracy on this subset by 4.2\% which is a significant boost. We believe this editing approach is a promising direction for future use, where practitioners could for example notice an incorrect prediction in a production system, quickly devise a fix for that prediction, and check how it affects other predictions using validation data.

\section{Conclusion}

We have presented \algoname{}, a fully automated and scalable method for generating concept bottleneck models and used it to create the first high performing CBM on ImageNet. In addition, we have demonstrated how our models are easier to understand in terms of both global decision rules and reasoning for individual predictions. Finally, we show how to use this understanding to manually edit weights to improve accuracy in a trained ImageNet model.

\newpage
\section*{Acknowledgements}
The authors would like to thank anonymous reviewers for valuable feedback to improve the manuscript. The authors also thank MIT-IBM Watson AI lab for support in this work. This work was done during T. Oikarinen's internship at MIT-IBM Watson AI Lab. T.-W. Weng is supported by National Science Foundation under Grant No. 2107189.

% \section*{Reproducibility}
% % We have described our algorithm in detail, including hyper-parameters and other details in sections 3 and 4. 
% Our code is available at \href{https://github.com/Trustworthy-ML-Lab/Label-free-CBM}{https://github.com/Trustworthy-ML-Lab/Label-free-CBM}.

% \lily{add appendix summary here}

\bibliography{iclr2023_conference}
\bibliographystyle{iclr2023_conference}

\newpage

\appendix

\section{Appendix}
\subsection{Appendix Overview}
\label{sec:app:overview}
In this section we provide a brief overview of the Appendix contents. The Appendix is mostly focused on Ablation studies identifying the importance and effectiveness of different components of our pipeline, as well additional Figures to showcase more examples of our results.

First in section \ref{sec:app:limitation} we discuss the limitations of our method, then in section \ref{sec:cos_cubed_performance} we show that our proposed \textit{cos cubed} similarity function performs well in describing neuron functionality. In section \ref{sec:filter_ablation} we perform an ablation study on the effect of different concept set filters we have proposed, and in Sec \ref{sec:filter_example} we discuss our reasoning behind each filter in detail through an example. In section \ref{sec:conceptnet_comparison} we show that our GPT-3 generated concept set helps us reach higher accuracy than previous methods for creating initial concept set. In Appendix \ref{sec:app:no_sparsity} we show the effects of removing the sparsity constraint on our models, leading to higher accuracy but reduced interpretability. In Appendix \ref{sec:concept_regen} we show our results are consistent despite inherent randomness in our pipeline. Next, in Appendix \ref{sec:app:model_editing_details} we provide a detailed explanation of the procedure we used for model editing described in Sec \ref{sec:editing_weights}. Finally in Sections \ref{sec:app:extra_figs}, \ref{sec:app:extra_weights} and \ref{sec:random_input_explanations} we provide additional figures for model editing, individual decision explanations and model weight visualizations respectively.

\textcolor{blue}{In Appendix B we provide results of our large-scale crowdsourced human evaluation, which confirms that LF-CBM is indeed more interpretable than standard models according to two metrics.}

\subsection{Limitations}
\label{sec:app:limitation}

While our model performs well in terms of both interpretability and accuracy, it still has some limitations. For example, using a model like GPT-3 to produce concepts is stochastic, and sometimes may fail to generate important concepts for detecting a certain class. However our model is quite robust to small changes in the concept set as shown in sections \ref{sec:filter_ablation} and \ref{sec:concept_regen}. In general, automatic ways of generating concept set may sometimes lack the required domain knowledge to include some important concepts, and would perhaps be best used in collaboration with a human expert.

Second, our model is aimed to extend Concept Bottleneck Models to large datasets where labeled concept data is not available. However it works best on domains where CLIP performs well and as such may not perform as well on small datasets that require specific domain knowledge and labeled concept data is available, such as medical datasets. For such datasets we believe it is still better to use a method that can effectively leverage the labels available such as original Concept Bottleneck Models \cite{koh2020concept} or P-CBM\cite{yuksekgonul2022post}.

\subsection{Dissection performance of cos cubed similarity function}

\label{sec:cos_cubed_performance}

In this section we discuss how good the \textit{cos cubed} similarity function we defined in section \ref{sec:method_CBL} is at describing neuron functionality, i.e. how well it performs as a similarity function for CLIP-Dissect. Specifically we measured its performance on describing the roles of final layer neurons on a ResNet-50 trained on ImageNet (where we know the ground truth role of the neurons) in terms of both accuracy and average distance from the correct description in a text embedding space. We followed the quantitative evaluation methodology of \citep{oikarinen2022clip}, which describes this process in more detail. We compared the results against using CLIP-Dissect with two different similarity functions, \textit{softWPMI} which was the best function identified by \citep{oikarinen2022clip} and \textit{cos} which is simple a differentiable baseline. We can see \textit{cos cubed} performs es well as \textit{softWPMI} in terms of similarity in text embedding space, while being slightly worse in terms of accuracy, and is much better than simple $cos$ similarity. This is impressive given \textit{cos cubed} is fully differentiable while \textit{SoftWPMI} is not differentiable. 

\begin{table}[h!]

\centering
\scalebox{0.85}{
\begin{tabular}{@{}ll|cccccc@{}}
\toprule
 &  & \multicolumn{1}{l}{} & \multicolumn{1}{l}{} & \multicolumn{1}{c}{$D_{probe}$} & \multicolumn{1}{l}{} & \multicolumn{1}{l}{} &  \multicolumn{1}{l}{} \\ \cmidrule{3-8}
\textbf{Metric} & \multicolumn{1}{c|}{\begin{tabular}[c]{@{}c@{}}\textbf{Similarity}\\ \textbf{function} \end{tabular}} & \multicolumn{1}{c}{\begin{tabular}[c]{@{}c@{}}CIFAR100\\  train\end{tabular}} & \multicolumn{1}{c}{\begin{tabular}[c]{@{}c@{}}Broden\\ {}\end{tabular}} & \multicolumn{1}{c}{\begin{tabular}[c]{@{}c@{}}ImageNet\\  val\end{tabular}} & \multicolumn{1}{c}{\begin{tabular}[c]{@{}c@{}}ImageNet val\\  + Broden\end{tabular}} & \multicolumn{1}{c}{\begin{tabular}[c]{@{}c@{}}Average\\ {} \end{tabular}} & \multicolumn{1}{c}{\begin{tabular}[c]{@{}c@{}}Differentiable\\ {} \end{tabular}} \\ \hline
mpnet & cos & 0.2761 & 0.215 & 0.2823 & 0.2584 & 0.2580 & \textbf{Yes}
\\
cos similarity & SoftWPMI & \textbf{0.3664} & 0.3945 & \textbf{0.5257} & \textbf{0.5233} & \textbf{0.4525} & No \\
& cos cubed & 0.3579 & \textbf{0.4101} & 0.5187 & 0.5223 & 0.4523 & \textbf{Yes} \\
  \midrule

Top1 accuracy & cos & 8.50\% & 5.70\% & 15.9\% & 11.4\% & 10.38 \% & \textbf{Yes}
\\
 & SoftWPMI & \textbf{46.20}\% & \textbf{70.50} \% & \textbf{95.00}\% & \textbf{95.40}\% & \textbf{76.78}\% & No \\ 
  & cos cubed & 31.00\% & 49.40 \% & 87.40\% & 85.50\% & 63.33\% & \textbf{Yes} \\
 \bottomrule
\end{tabular}
}
\caption{Comparison of the performance between similarity functions. We look at the final layer of ResNet-50 trained on ImageNet. We use 20,000 most common English words as the concept set for mpnet cos similarity and ImageNet classes as the concept set for top1 accuracy. We can see \textit{cos cubed} performs much better than simple \textit{cos}, and almost as well as \textit{SoftWPMI}.}
\label{tab:distance_comparison}
\end{table}

\newpage

\subsection{Ablation: Effect of concept filters}

\label{sec:filter_ablation}

In this section we study how each step in our proposed concept filtering effects the results of our method. In general our use of filters has two main aims:
\begin{itemize}
    \item To improve the interpretability of our models
    \item Improve computational efficiency and complexity by reducing the number of concepts
\end{itemize}

The filters are not designed to improve performance of the models in terms of accuracy, in fact we would expect them to slightly reduce accuracy as a model with more concepts is generally larger and more powerful. 

To evaluate the effect each individual filter has we trained models on both CIFAR10 and ImageNet while deactivating one filter at a time, as well as one without using any filters at all. The results are displayed in Table \ref{tab:filter_ablation}. We can see that the accuracy of our models is not at all sensitive to the choice of filters, with ImageNet accuracy remaining constant and CIFAR10 slightly increasing its accuracy with less filters. We were unable to train an ImageNet model without any filters as training with the large number of concepts required more memory than we had available in our system. Additionally the ImageNet model without similarity to other concepts filter had worse accuracy and less sparsity than other tested models, which we think may be caused by the GLM-SAGA optimizer not finding as good of a solution when the number of concepts increased too much.

\begin{table}[h!]
\centering
\begin{tabular}{@{}lrrrr@{}}
\toprule
Filters & \multicolumn{1}{l}{\begin{tabular}[c]{@{}l@{}}CIFAR10\\  Accuracy\end{tabular}} & \multicolumn{1}{l}{\begin{tabular}[c]{@{}l@{}}CIFAR10\\  \#concepts\end{tabular}} & \multicolumn{1}{l}{\begin{tabular}[c]{@{}l@{}}ImageNet\\  Accuracy\end{tabular}} & \multicolumn{1}{l}{\begin{tabular}[c]{@{}l@{}}ImageNet\\  \#concepts\end{tabular}} \\ \midrule
All filters & 86.26\% & 142 & 71.89\% & 4380 \\
No length filter & 86.55\% & 146 & 71.88\% & 5347 \\
No similarity to classes filter & 86.74\% & 148 & 71.93\% & 4671 \\
No similarity to other concepts filter & 86.42\% & 151 & 70.90\%* & 6515 \\
No CLIP activation filter & 86.41\% & 147 & 71.92\% & 4478 \\
No projection accuracy filter & 86.33\% & 147 & 71.88\% & 4462 \\
No filters at all & 86.56\% & 177 & N/A & 9087 \\ \bottomrule
\end{tabular}
\caption{Effect of our individual concept filters on the final accuracy and number of concepts used by our models. *We were unable to train the model to be sparse enough, results from a less sparse model.}
\label{tab:filter_ablation}
\end{table}

\newpage

\subsection{Further discussion and example of concept filtering}

\label{sec:filter_example}

To further show the effects of each filter, we will showcase the full filtering procedure for our CIFAR10 model with all filters. We start with a freshly generated initial concept set from GPT-3, which has a total of 177 concepts. We then take the following steps to filter down concepts:

\begin{enumerate}
    \item \textbf{Delete concepts that are too long.} This step is important to make sure concepts are easy to visualize and simple non-convoluted concepts. For CIFAR-10 this leads to deletion of 3 concepts: \newline
    - \textit{white spots on the fur (in some cases)}, \textit{feline features (e.g., whiskers, ears)} and \textit{legs with two toes pointing forward and two toes pointing backwar}. \newline
    174 concepts remain.
    \item \textbf{Delete concepts too similar to output classes.} This step is required for the explanations to be informative, as it helps avoid trivial explanations such as: "This is image is a cat because it is a cat". This step removes the following CIFAR-10 concepts (with the class name it is too close to shown in brackets): \newline 
    - a cat (cat), a deer (deer), a horse (horse), a plane (airplane), a car (automobile), car (automobile), cars (automobile), vehicle (automobile), animal (dog), truck driver (truck)
    \newline
    164 concepts remain
    \item \textbf{Delete concepts too similar to other concepts.} In this step we aim to delete duplicate concepts from the concept set, i.e. two concepts that have the same semantic meaning should not both be part of the concept set. Duplicate concepts will cause unnecessary computational cost and confusing explanations. For CIFAR-10 we delete the following concepts (with the concept it is too similar to in brackets): \newline
    - 4 wheels (four wheels), a food bowl (a bowl), a furry, four-legged animal (a large, four-legged mammal), a gasoline station (a gas station), a large boxy body (a large body), a large, muscular body (a large body), a street (a road), a strong engine (an engine), large, bulging eyes (large eyes), the ocean (the sea) \newline
    154 concepts remain
    \item \textbf{Remove concepts not present in training data.} In this step we remove all concepts that CLIP thinks are not really present in training data. The purpose is to avoid learning neurons that don't correctly represent their target concepts, as concepts missing from the dataset are unlikely to be learned correctly in the Concept Bottleneck Layer. For CIFAR-10 this step deletes the following concepts:\newline
    - a bit, a mechanic, legs, long legs, passengers, several seats inside, windows all around \newline
    147 concepts remain
    \item \textbf{Remove concepts we can't project accurately.} The purpose of this step is similar to the previous step, we want to remove all concepts that are not faithfully represented by the CBL. To do this we evaluate the similarity score between the target concept and the activations of our new neuron using CLIP-Dissect \cite{oikarinen2022clip} on the validation data, and delete concepts with similarity less than 0.45 which was determined to be a good indicator of faithful concept representation. For CIFAR-10 we delete the following concepts: \newline 
    - a bed, a coffee mug, a crew, a house, a wheel \newline
    We are left with our 142 final concepts.

\end{enumerate}

In total our method has 5 different cutoff parameters that can be tuned. Since the main purpose of these filters is to make the network more interpretable, their values were mainly chosen through trial and error to find ones that produce the most useful and explainable models. Most of these cutoffs are independent of the dataset chosen, and the only one we changed from one dataset to another was the value of cutoff for images being present in the dataset (filter 4). We used a cutoff of 0.25 for CIFAR-10 and CIFAR-100, 0.26 for CUB-200 and 0.28 for Places and ImageNet. We found we had to change this cutoff as CLIP doesn't activate very highly on low resolution images of CIFAR, and in general we get higher top5 activations on larger datasets. Since the other 4 cutoffs are fixed, using our method for a new dataset won't require a large hyperparameter search, and as seen in Table \ref{tab:filter_ablation} the overall performance is not very sensitive to specific filters.

\begin{figure}[h!]
    \centering
    \includegraphics[width=0.95\linewidth]{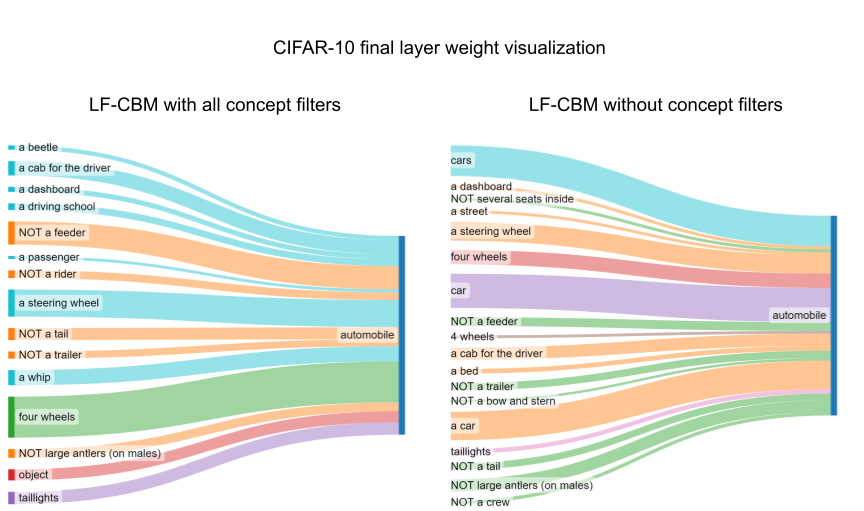}
    \caption{\textit{Comparison of the final layer weights between CIFAR-10 model trained with and without filters.}}
    \label{fig:filter_ablation}
\end{figure}

To visualize the effect of our filters, in Figure \ref{fig:filter_ablation} we visualize the final layer weights for CIFAR-10 class "automobile" trained with and without filters. We can see the model with filters has found a reasonable decision rule, with largest weights corresponding to \textit{four wheels} and \textit{a steering wheel}. On the other hand, the weights for the model without filters are quite problematic. First, the 3 largest weights are \textit{cars}, \textit{car} and \textit{a car}. Not only do they have the same meaning as the final class itself (filter 2), thus lacking any explanatory power, they are duplicates of each other (filter 3), making the explanation unnecessarily compilicated. Finally we see the unrelated concept \textit{bed} which is not present in CIFAR-10 having a small positive weight, which should be removed by either filter 4 or 5.

\subsection{\rebuttal{Ablation: Effect of Initial Concept set}}

\label{sec:conceptnet_comparison}

\rebuttal{In this section we evaluate the effect our initial concept set generator has on our results. In our paper we proposed a new method of generating descriptions using GPT-3 \cite{gpt3}, and in this section we compare those results to the case where we generate the initial concept set using ConceptNet \cite{conceptnet} as proposed by \cite{yuksekgonul2022post}. In general ConceptNet only works well when prompted with relatively common single word phrases, but many of class names for CUB200, Places365 and ImageNet consist of multiple words, such as the ImageNet class \textit{great white shark}. To overcome this issue we split multi-word class names into single word components and add concepts related to each word to the concept set, in this case \textit{great}, \textit{white} and \textit{shark}. After creating the initial concept set we use the same filtering procedure described in section \ref{sec:concept_set}B for both models. Table \ref{tab:conceptset} displays the accuracy comparison between models trained on different initial concept setst. In general using GPT3 seems to provide a small (0.1-1.5\%) accuracy boost on all datasets, with the exception of CUB200 where the ConceptNet model fails completely. This is because ConceptNet is unable to generate concepts for the highly specific class names in CUB200, such as \textit{Groove billed Ani}, while GPT-3 is not troubled by this.}

\begin{table}[h!]
\centering
\begin{tabular}{@{}lccccc@{}}
\toprule
Model\textbackslash{}Dataset & \multicolumn{1}{l}{CIFAR10} & \multicolumn{1}{l}{CIFAR100} & \multicolumn{1}{l}{CUB200} & \multicolumn{1}{l}{Places365} & \multicolumn{1}{l}{ImageNet} \\ \midrule
LF-CBM (GPT-3 [original]) & \textbf{86.40\%} & \textbf{65.13\%} & \textbf{74.31\%} & \textbf{43.68\%} & \textbf{71.95\%} \\
LF-CBM (ConceptNet) & 86.30\% & 63.62\% & 2.19\% & 42.49\% & 71.52\% \\ \bottomrule
\end{tabular}
\caption{\rebuttal{Accuracy comparison between using different methods for generating initial concept set.}}
\label{tab:conceptset}
\end{table}

\clearpage 
\newpage

\subsection{Ablation: LF-CBM without sparsity constraint}

\label{sec:app:no_sparsity}

\begin{figure}[h!]
    \centering
    \includegraphics[width=0.95\linewidth]{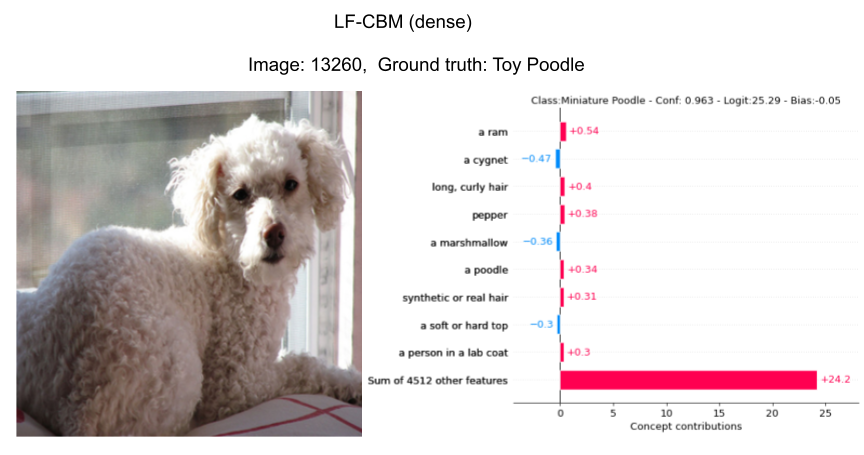}
    \caption{Dense CBM explanation for the same image as the fourth image in Figure \ref{fig:imagenet_samples}. We can see the dense model is practically uninterpretable.}
    \label{fig:dense_cbm_explanation}
\end{figure}

To estimate the effect sparsity has on our models, we show the accuracy of LF-CBM models trained without sparsity constraints (LF-CBM (dense)) in Table \ref{tab:dense_cbm}. We can see that not using a sparsity constraint greatly improves accuracy (except for CUB200 which starts to overfit), but it comes at a very large cost for interpretability as can be seen in Figure \ref{fig:dense_cbm_explanation}, where almost none of the decision is explained by the 10 most highly contributing concepts.

\begin{table}[h!]
\centering
\begin{tabular}{lrrrrl}
\hline
Model\textbackslash{}Dataset & \multicolumn{1}{l}{CIFAR10} & \multicolumn{1}{l}{CIFAR100} & \multicolumn{1}{l}{CUB200} & \multicolumn{1}{l}{Places365} & ImageNet \\ \hline
LF-CBM (sparse [original]) & 86.40\% & 65.13\% & 74.31\% & 43.68\% & 71.95\% \\
LF-CBM (dense) & 87.50\% & 67.93\% & 74.25\% & 48.25\% & 74.09\% \\ \hline
\end{tabular}
\caption{\rebuttal{Results of training our CBM without sparisity constraints on the final layer.}}
\label{tab:dense_cbm}
\end{table}

\newpage

\subsection{\rebuttal{Consistency of concept set generation}}

\label{sec:concept_regen}

\begin{figure}[h!]
    \centering
    \includegraphics[width=0.9\linewidth]{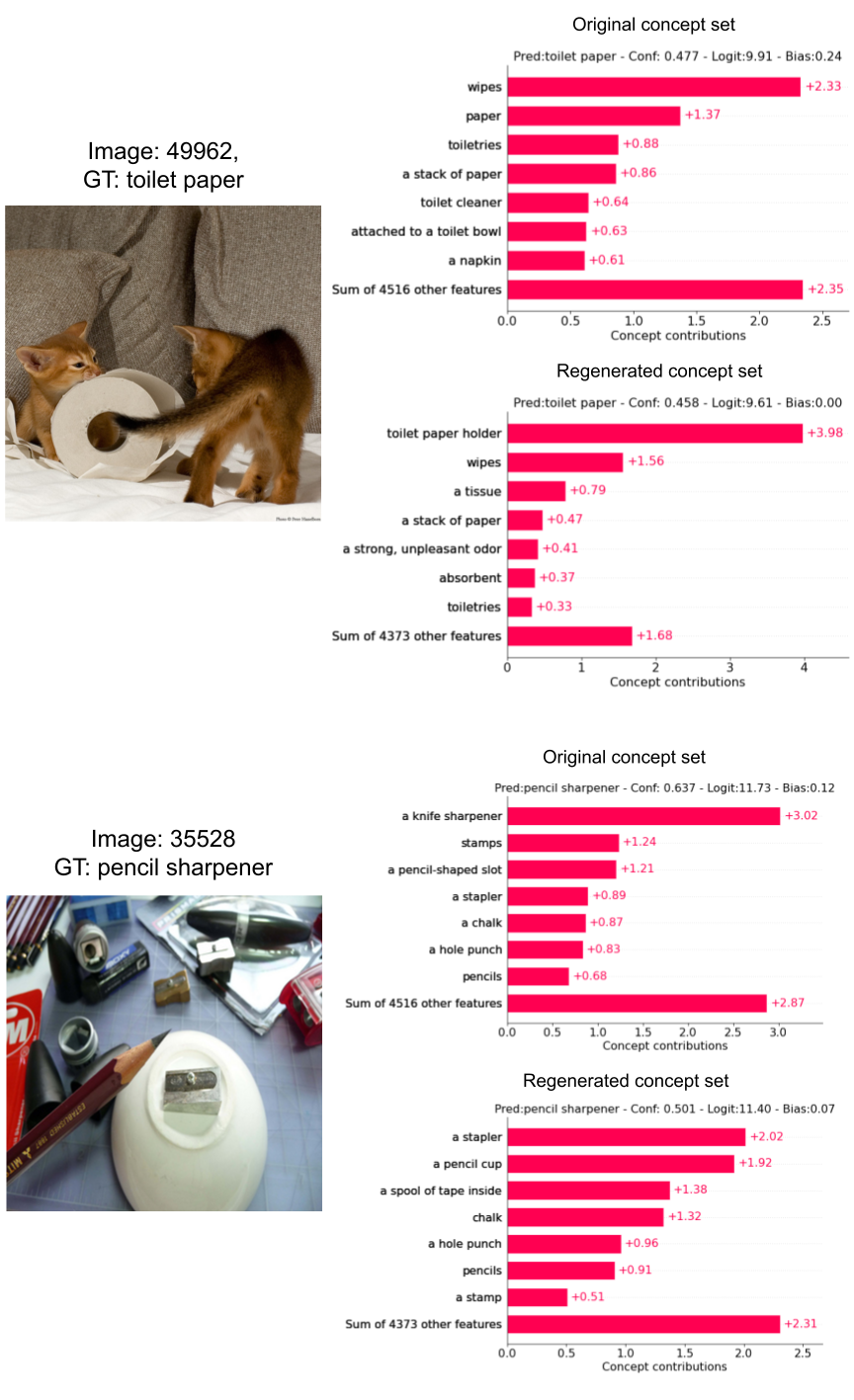}
    \caption{Difference in concepts used for a decision and their weights for models trained with original vs regenerated concept set, shown on two random images.}
    \label{fig:conceptset_consistency}
\end{figure}

\rebuttal{Since the initial concept set is generated using GPT-3, it is a random process and in this section we explore how much this noise effects our final results. When regenerating initial concept set and running our concept filtering pipeline for our ImageNet model, we got 4380 concepts, compared to the original 4523 concepts. In terms of accuracy, the model learned with regenerated concept set reached an accuracy of 71.89\%, compared to the average of 71.95\% with the original concept set. This indicates that the difference is very small in terms of accuracy. However, the exact concepts used and model weights can be pretty different as shown in Figure \ref{fig:conceptset_consistency}, despite all being relevent (e.g. there are many concepts that appear on both concept sets, e.g. Image 49962 has important concepts of "wipes", "toiletries" on both, though they have different corresponding contributions).  Figure \ref{fig:conceptset_consistency} shows the explanations for decisions on two random images for the original model and model trained with regenerated concept set.}

\clearpage
\newpage

\subsection{More details on Model editing procedure}

\label{sec:app:model_editing_details}

Here we provide an improved explanation of Sec 5.2 on how we can edit the final layer weights of our proposed Label-Free CBM to correct incorrect predictions. Our procedure includes 3 steps:

\textbf{Step 1: Find an input image $x_i$ where the model makes a Type 4 Error (incorrect final layer weight)}:

This is done by visualizing incorrect model predictions and their explanations similar to Figure \ref{fig:contributions}. To identify if it is a Type 4 error, first we check that the ground truth label of the input image is correct and unambiguous (therefore not type 1 error described in sec 5.1). Next, we check that the highly activating concepts look correct (therefore not type 2 or type 3 error described in sec 5.1).

\textbf{Step 2: Identify a concept to edit}:

We start by listing the highest activating concepts for the chosen input, i.e. the highest elements of $f_c(x_i)$ (after normalization). This allows us to identify concepts that are the most important for this specific image while not too important for other images. From these concepts we use our domain knowledge (and/or internet search) to understand which concepts are relevant to making this decision. 

For example, the image in the right panel of Figure \ref{fig:model_editsl} is originally predicted as a “Shopping basket”, while the ground truth is a “Hamper”. The 5 most highly activating concepts for this image are: “a basket”, “made of rope or string”, “a rope”, “a laundry basket” and “a fishing net”. After a short investigation we find that although the classes “Hamper” and “Shopping basket” are very similar, hampers are more often constructed in a woven like manner. Therefore, we identify the concept of “made of rope or string” as relevant for capturing this difference, and choose it for editing in the next step.

\textbf{Step 3: Change weights by sufficient magnitude}:

Once we have selected the concept to be corrected, we will compute how much the magnitude of the associated final layer weight should be changed (denoted as $\Delta w$, $\Delta w \in \mathbb{R}$) and then edit the weights in the following way:
\begin{equation}
    W_{F[gt,concept]} \leftarrow W_{F[gt,concept]}+\Delta w, \;\; W_{F[pred,concept]} \leftarrow W_{F[pred,concept]} - \Delta w
\end{equation}
Note that $W_{F[i,j]}$ denotes the $(i,j)$ element in the matrix $W_F$. The goal of editing is to correct the inaccurate prediction on this specific instance while minimizing effect on other predictions. To calculate $\Delta w$ needed to flip prediction, we first calculate the difference in logits $\Delta a$ (before softmax) before the edit, $\Delta a = W_{F[pred,:]} f_c(x_i) - W_{F[gt,:]} f_c(x_i)$. 

Since we would like the prediction to be corrected to gt class, our goal is to design $\Delta w$ such that
\begin{equation}
    (W_{F[gt,:]}+\Delta w \cdot e) f_c(x_i) - (W_{F[pred,:]}-\Delta w \cdot e) f_c(x_i) > 0
\end{equation}
where $e$ is a one-hot row vector with the entry $e_{[concept]} = 1$. Let $b = (W_{F[gt,:]}+\Delta w \cdot e) f_c(x_i) - (W_{F[pred,:]} - \Delta w \cdot e) f_c(x_i)$, where $b$ is a nonnegative constant deciding how large of a margin we want the correct prediction to have. Since $\Delta a = W_{F[pred,:]} f_c(x_i) - W_{F[gt,:]} f_c(x_i)$, we have: 
\begin{equation}
    2 \Delta w \cdot f_c(x_i)_{[concept]}- \Delta a = b \Rightarrow \Delta w =  (\Delta a+b)/(2 f_c(x_i)_{[concept]})
\end{equation}
Typically we use values for $b$ between 0.2 and 2 to calculate required $\Delta w$. 

It is worth noting that when the model weights are edited, it might affect other image’s predictions too due to the change of weight parameters. Thus, with the goal to correct the wrong predictions while not affecting other already correct predictions, we suggest each edit should be checked with a validation dataset before applying on a target model. 
As described in the end of Sec 5 in the manuscript, with the above proposed model weight editing, we are able to improve the overall model accuracy from 71.98\% to 72.02\% by performing the model editing for only 5 different images, which is a non-negligible improvement. Since the edits only affect weights for 10/1000 classes, this corresponds to a 4\% accuracy boost on the affected classes.

\clearpage 
\newpage

\subsection{Additional figures}

\label{sec:app:extra_figs}

\begin{figure}[h!]
    \centering
    \includegraphics[width=0.95\linewidth]{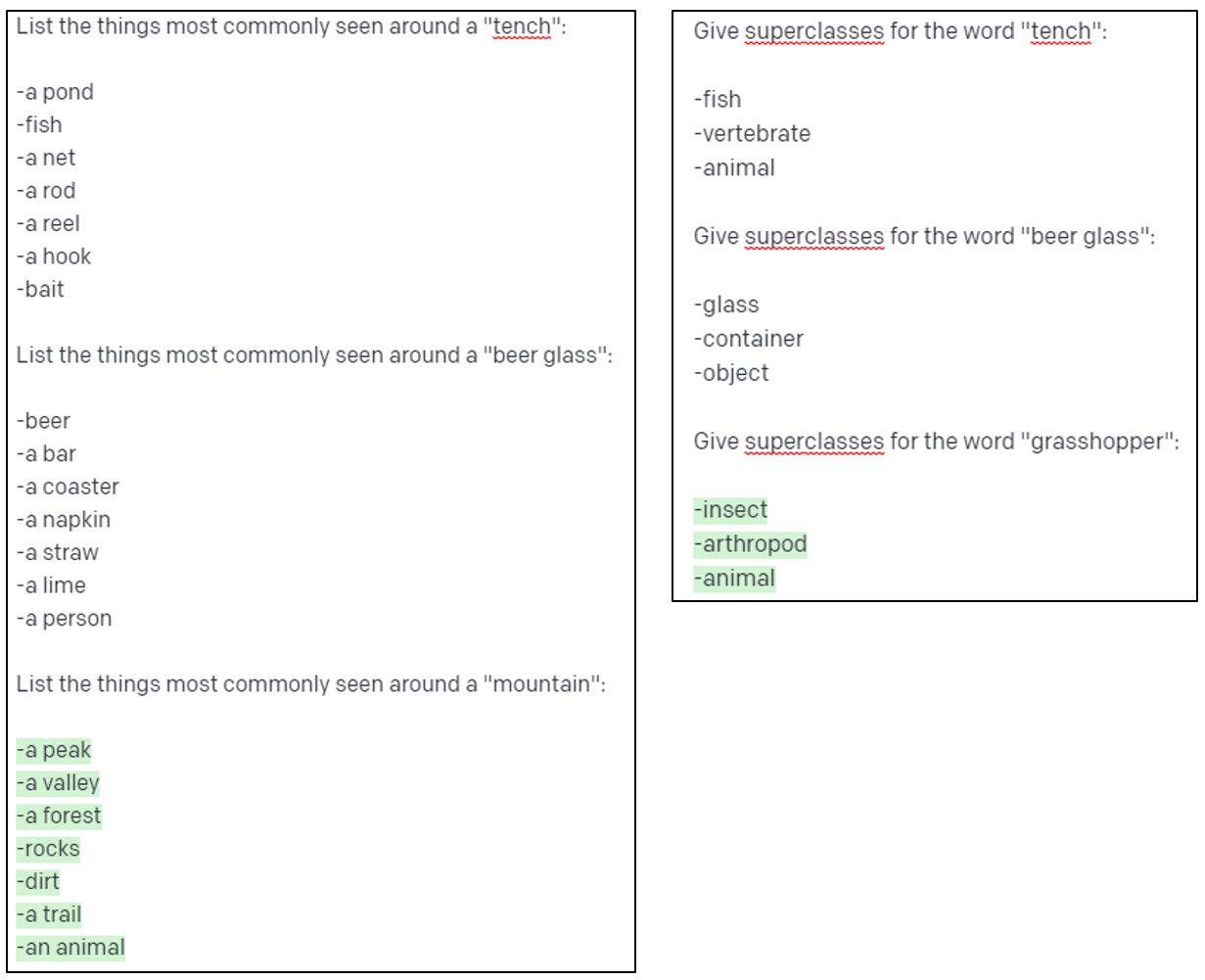}
    \caption{Full prompts used for GPT-3 concept set creation. Text in green generated by GPT, rest is our prompt.}
    \label{fig:gpt_1}
\end{figure}

\begin{figure}
    \centering
    \includegraphics[width=0.95\linewidth]{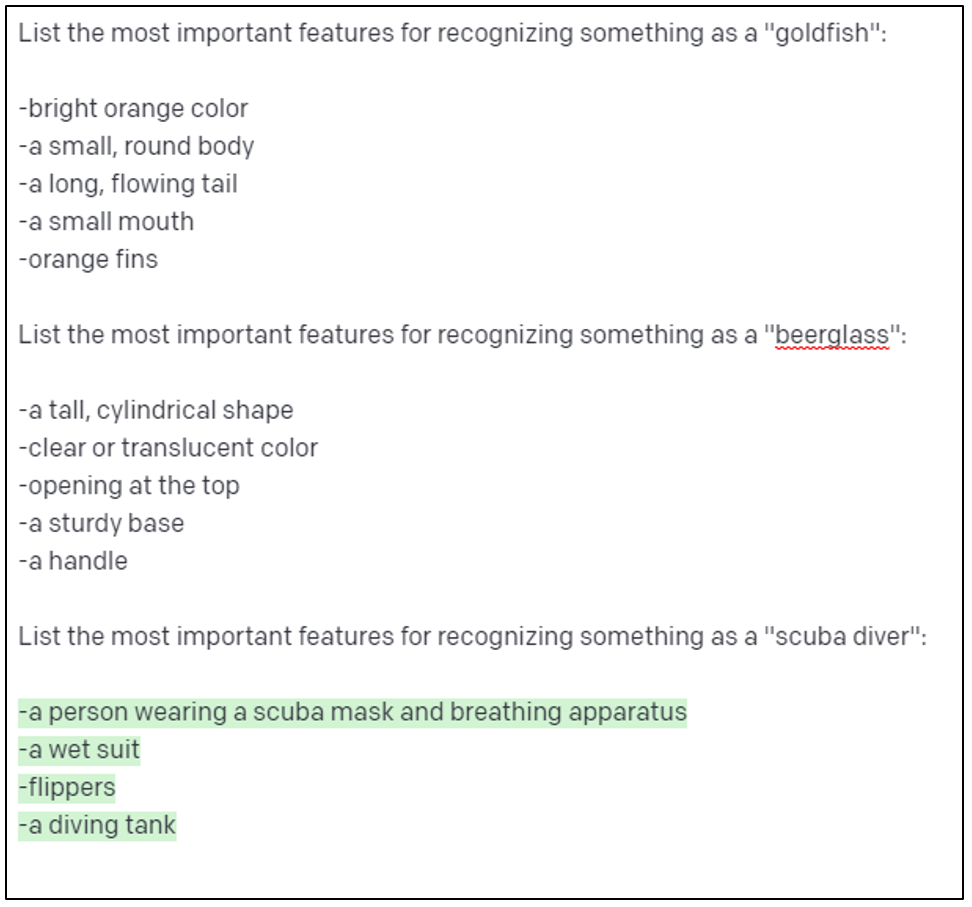}
    \caption{Full prompts used for GPT-3 concept set creation. Text in green generated by GPT, rest is our prompt.}
    \label{fig:gpt_2}
\end{figure}

Figures \ref{fig:gpt_1}, \ref{fig:gpt_2} provide examples of the full prompts we used for GPT-3, as well as GPT outputs. For all experiments we used the text-davinci-002 model available through OpenAI API.

\begin{figure}
    \centering
    \includegraphics[width=0.9\linewidth]{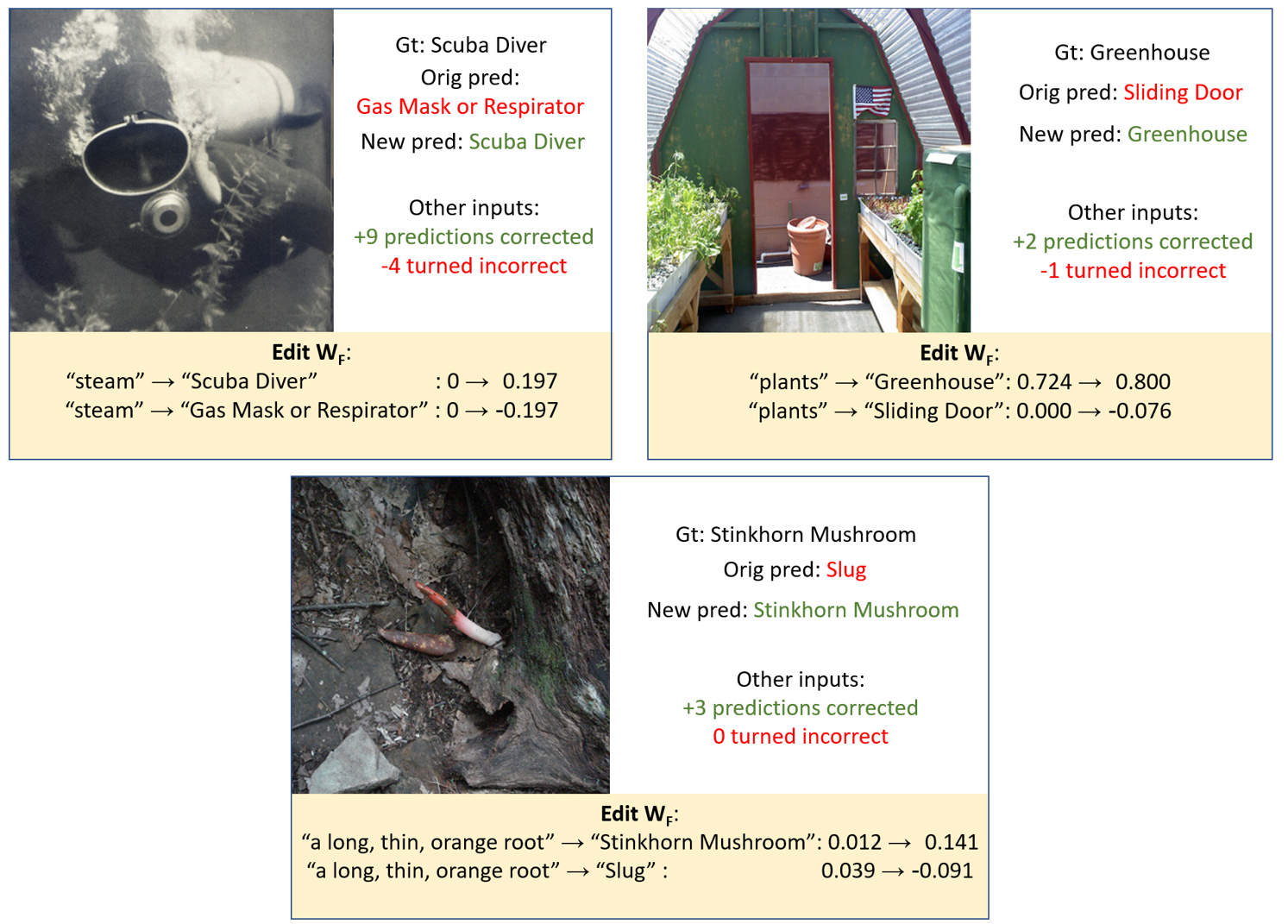}
    \caption{The other 3 model edits we performed in our experiment}
    \label{fig:appendix_edits}
\end{figure}

\begin{figure}
    \centering
    \includegraphics[width=0.9\linewidth]{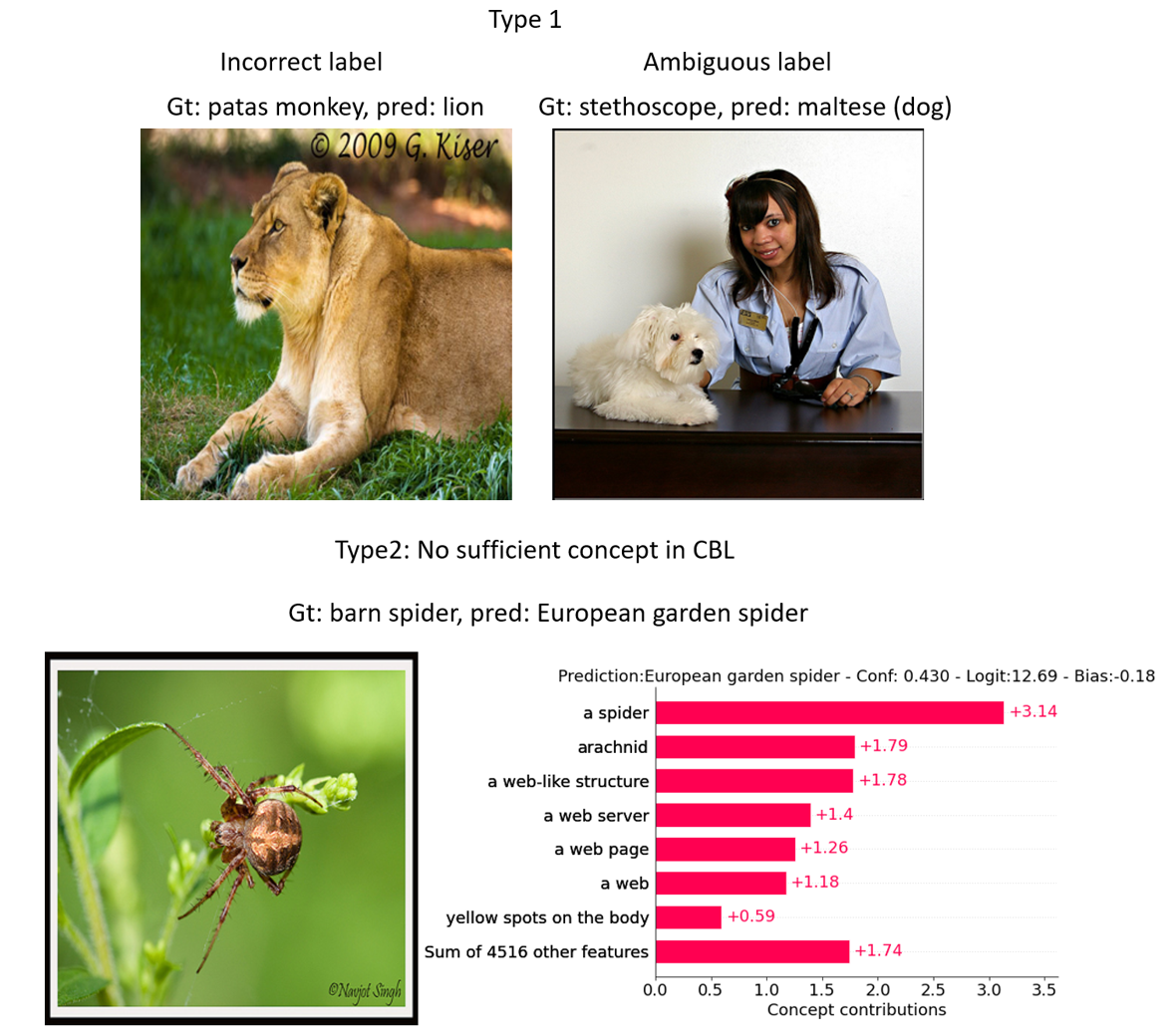}
    \caption{Examples of Type 1 and type 2 model errors.}
    \label{fig:error_types}
\end{figure}

Figure \ref{fig:appendix_edits} shows the additional model edits performed in our ImageNet CBM experiment, and Figure \ref{fig:error_types} showcases Type 1 and Type 2 errors made by our ImageNet CBM.

\clearpage 
\newpage

\subsection{Additional weight visualizations for random classes}

\label{sec:app:extra_weights}

\begin{figure}[h!]
    \centering
    \includegraphics[width=0.95\linewidth]{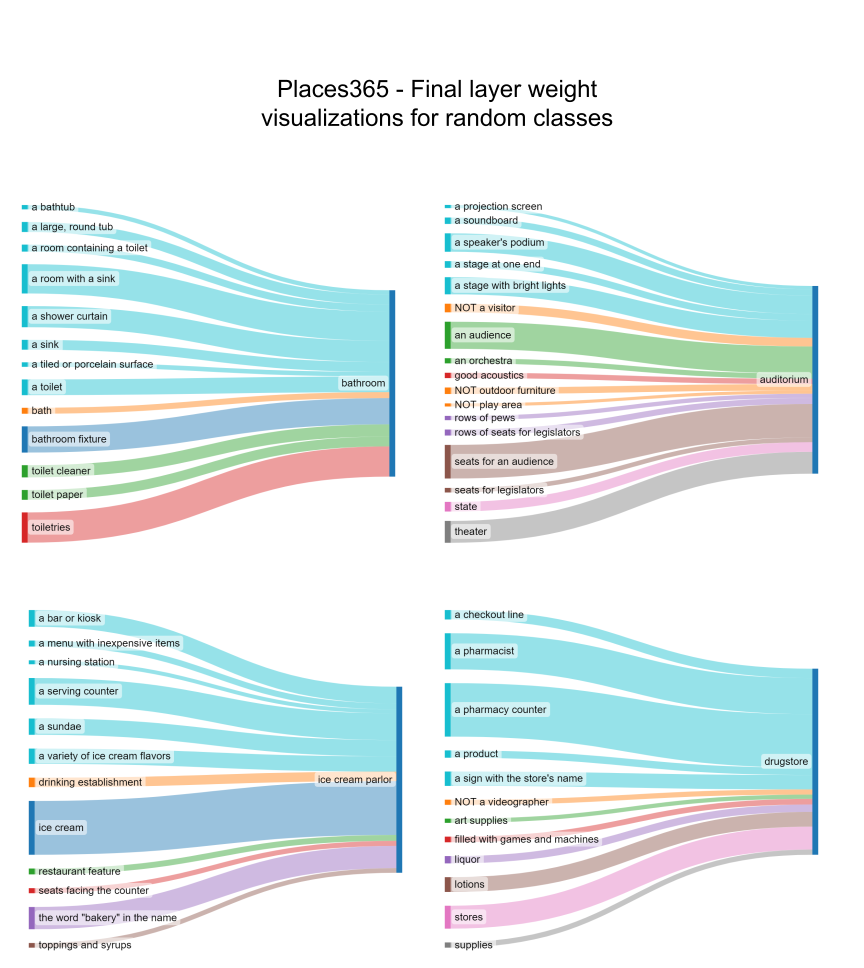}
    \caption{Weight visualizations for 4 randomly chosen output classes for our CBM trained on Places365.}
    \label{fig:places_weights}
\end{figure}

\begin{figure}[h!]
    \centering
    \includegraphics[width=0.9\linewidth]{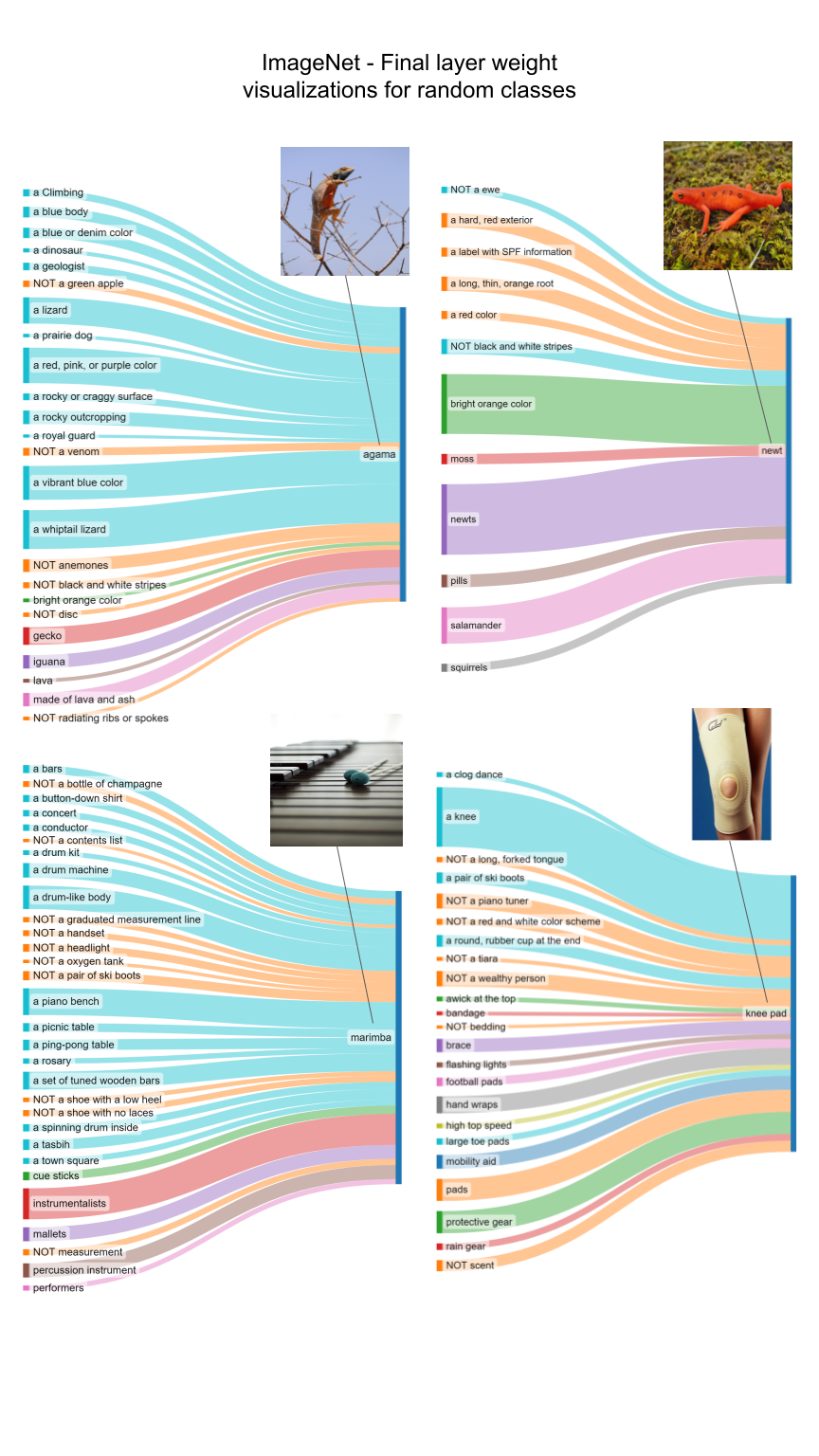}
    \caption{Weight visualizations for 4 randomly chosen output classes for our CBM trained on ImageNet, as well as an example image of each class to clarify their meaning.}
    \label{fig:imagenet_weights}
\end{figure}

In figures \ref{fig:places_weights} and \ref{fig:imagenet_weights} we have visualized the final layer weights leading to 4 randomly chosen output classes for both Places365 and ImageNet. These were created using the same procedure as Figure \ref{fig:weights}, with minor differences such as focus on single class at a time and no manual editing of concept colors according to semantic similarity. We can see that the concepts that have large weights are indeed relevant to the class, and overall the weights look reasonable.

\clearpage
\newpage

\subsection{Additional explanations for random images}

\label{sec:random_input_explanations}

\begin{figure}[h!]
    \centering
    \includegraphics[width=0.9\linewidth]{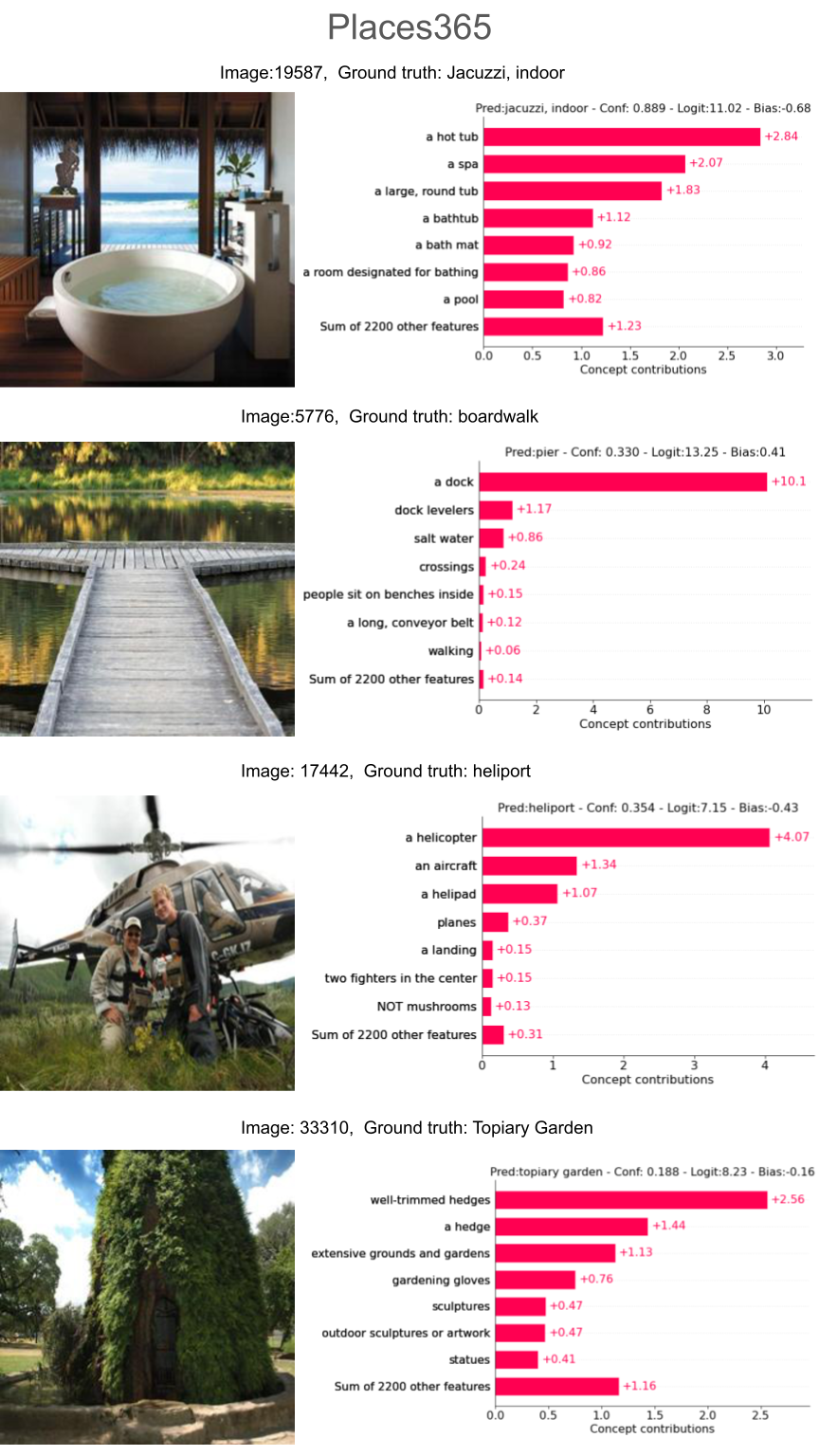}
    \caption{Explanations for 4 randomly chosen input images for our CBM trained on Places365.}
    \label{fig:places_samples}
\end{figure}

\begin{figure}[h!]
    \centering
    \includegraphics[width=0.9\linewidth]{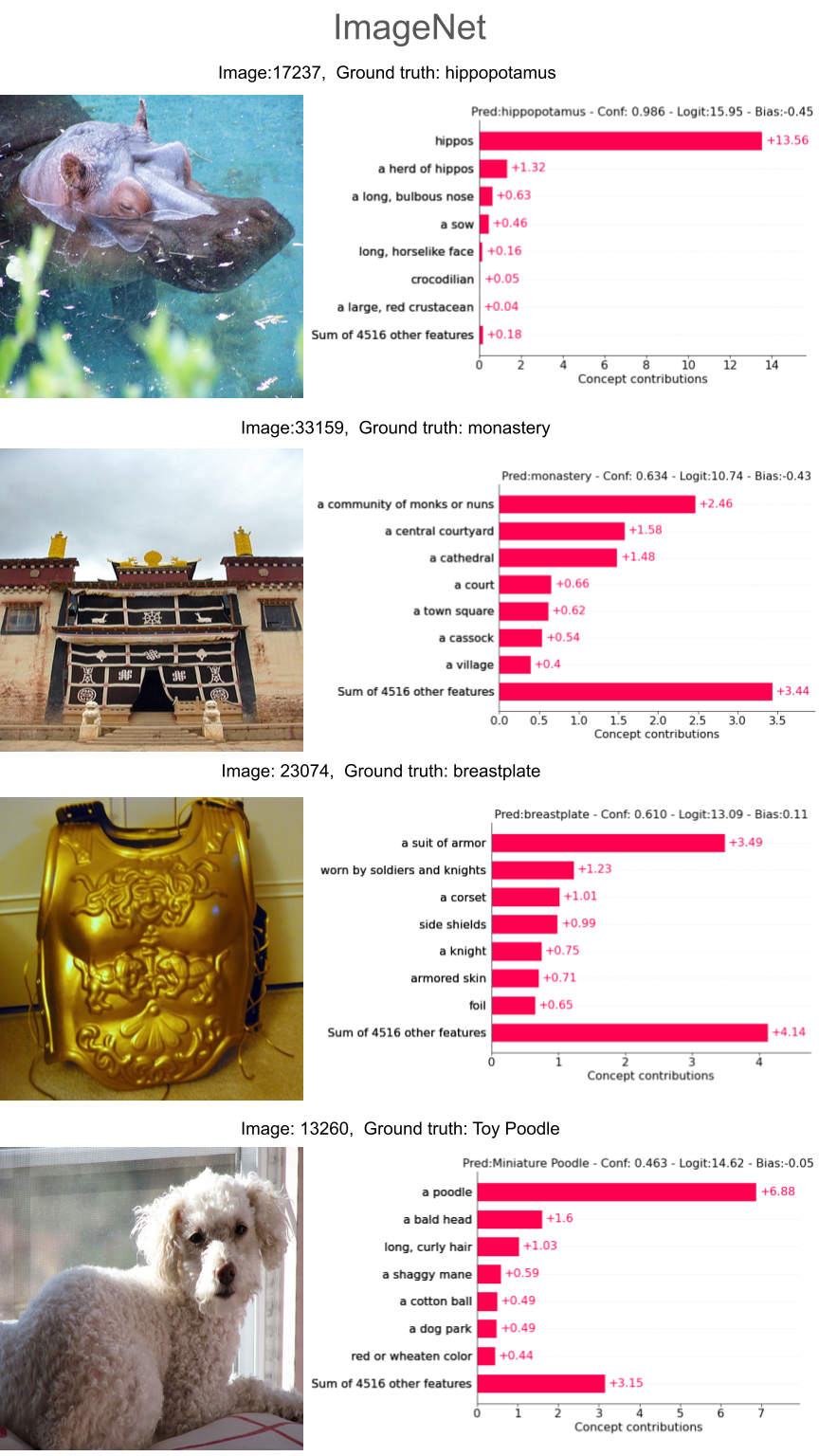}
    \caption{Explanations for 4 randomly chosen input images for our CBM trained on ImageNet.}
    \label{fig:imagenet_samples}
\end{figure}

\begin{figure}[h!]
    \centering
    \includegraphics[width=0.9\linewidth]{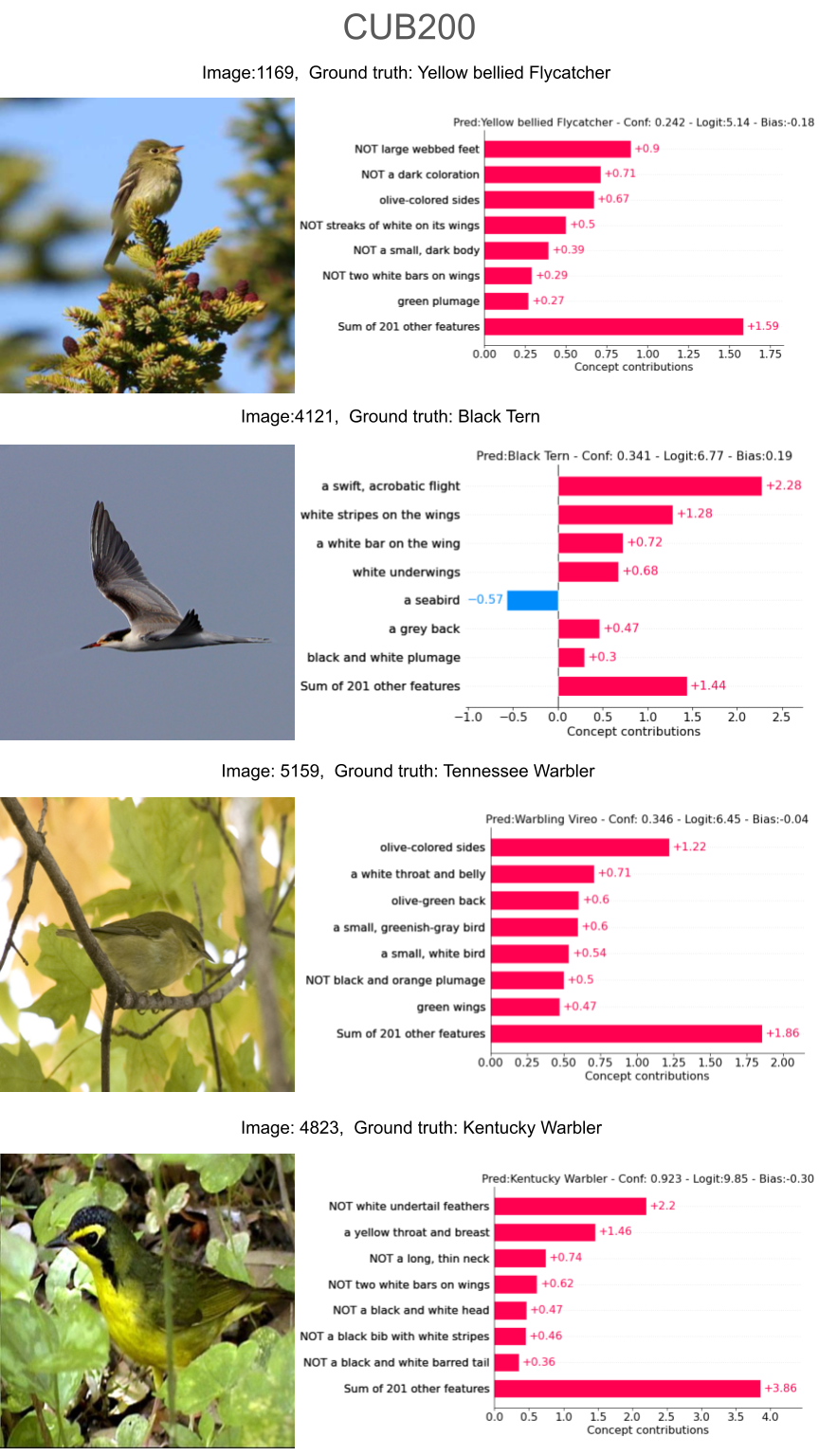}
    \caption{Explanations for 4 randomly chosen input images for our CBM trained on CUB-200.}
    \label{fig:cub_samples}
\end{figure}

In figures \ref{fig:places_samples}, \ref{fig:imagenet_samples}, \ref{fig:cub_samples} we display additional samples of decision explanations for 4 randomly chosen input images for each of Places365, ImageNet and CUB-200 respectively. Overall, it can be seen that the concepts and explanations are of good quality, even with the CUB-200 dataset which may require very specific knowledge and concepts to create CBMs. This result demonstrates the effectiveness of our LF-CBM, which does not use the expert-knowledge concept sets unlike existing CBMs.

\clearpage
\newpage

\section{User Study on the Interpretability of LF-CBM}
\label{sec:user_study}

\textcolor{blue}{Note: The results in this section were added after the ICLR camera ready deadline to further support the interpretability of our proposed method and are not included in the official ICLR paper.}

To get a quantitative measure of the interpretability of our models, we conducted a large scale study on Amazon Mechanical Turk. We evaluated our models on two tasks:
\begin{enumerate}
    \item Evaluating how well the neurons in CBL correspond to their target concepts
    \item Evaluating how reasonable the explanations for the decisions of our models are
\end{enumerate}

\subsection{Task 1: Do neurons in CBL correspond to their target concept?}

For our explanations to be faithful, it is very important that the neurons we learned in the Concept Bottleneck Layer (Section \ref{sec:method_CBL}) indeed correspond to their target concepts. 

\begin{figure}[b!]
    \centering    \includegraphics[width=0.95\textwidth]{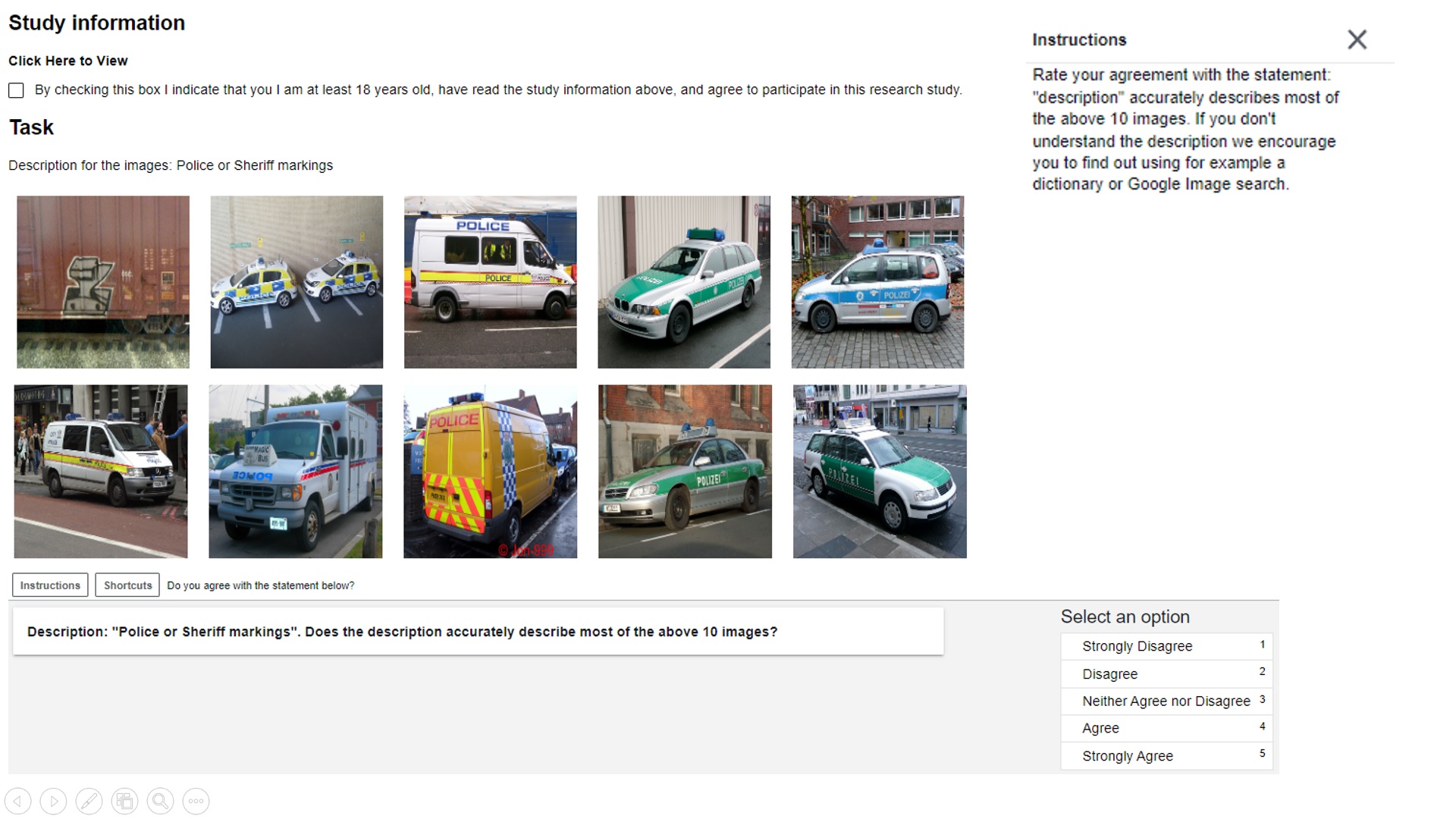}
    \caption{User Interface shown to Mechanical Turk users for Task 1.}
    \label{fig:task1_interface}
\end{figure}

\textbf{Setup:} To test this, we conducted a large scale evaluation on Amazon Mechanical Turk, where we evaluated all 4505 neurons in the CBL of our trained ImageNet LF-CBM, with 3 ratings per neuron. Our experiment interface is shown in Figure \ref{fig:task1_interface}. The basic idea is to show users the 10 most highly activating images (from the validation dataset) for a neuron, and ask them to rate whether the description accurately matches the images on a 5-point scale, where 1=Strongly Disagree and 5=Strongly Agree. The description for a CBL neuron is its target concept.

As a baseline, we compared against neurons in the second to last layer of ResNet-50 trained on ImageNet, which was the backbone model used for ImageNet LF-CBM. Since these neurons don't originally have an associated concept, we used CLIP-Dissect~\citep{oikarinen2022clip} to generate explanations for each neuron, and used those as the description. We evaluated each of the 2048 neurons in the layer with 3 raters each. For CLIP-Dissect we used original SoftWPMI similarity function and ImageNet validation data as probing dataset.

We used raters in the United States, with greater than 98\% approval rating and at least 10000 previously approved HITs. Users were paid \$0.05 per task. Our experiment was deemed exempt from IRB approval by the UCSD IRB Board.

\begin{table}[]
\centering
\scalebox{0.8}{
\begin{tabular}{@{}lcccccc@{}}
\toprule
 & Avg score: & \begin{tabular}[c]{@{}l@{}}1 - Strongly Disagree\end{tabular} & 2 - Disagree & \begin{tabular}[c]{@{}l@{}}3 - Neither Agree \\or  Disagree\end{tabular} & 4 - Agree & \begin{tabular}[c]{@{}l@{}}5 - Strongly  Agree\end{tabular} \\ \midrule
LF-CBM & \textbf{3.91 $\pm$ 0.01} & 4.3\% & 9.2\% & 10.1\% & 43.6\% & 32.8\% \\
ResNet-50 & 3.65 $\pm$ 0.02 & 6.2\% & 13.9\% & 14.7\% & 39.2\% & 26.0\% \\ \bottomrule
\end{tabular}
}
\caption{Mechanical Turk results of Task 1. We can see the CBL neurons in LF-CBM were deemed to be significantly more interpretable than neurons in its standard backbone network, on average reaching "Agree" scores. Avg score indicates the average rating across all neurons and all evaluations, reported with standard error of the mean.}
\label{tab:mturk_task1}
\end{table}

\textbf{Results:} The results of this experiment are presented in Table \ref{tab:mturk_task1}. We can see that on average, the neurons in CBL are more interpretable than the neurons of a standard neural network, reaching average score of 3.91 vs 3.65. Since we evaluated thousands of neurons this result is highly statistically significant. For more intuitive numbers, 76.4\% of raters gave CBL neurons a rating of "Agree" or higher, while this number was only 65.2\% for standard neurons. Finally, to evaluate how reliable these crowdsourced results are, we had 4 authors perform the same experiment in a blind setting, each evaluating 100 neurons for both models. Overall the results were similar, we noticed authors were more critical of both descriptions but both sets of evaluators found LF-CBM neurons more interpretable.  The average author score was 3.66 for LF-CBM neurons, and 3.30 for neurons in ResNet-50. MTurk average scores for this subset were LF-CBM:3.93, ResNet-50: 3.68. 

\subsection{Task 2: How reasonable are our explanations for a single decision?}

In the second task, we wanted to evaluate whether users consider the explanations our LF-CBM produces for individual decisions to be reasonable.

\begin{figure}[t!]
    \centering
    \includegraphics[width=0.95\textwidth]{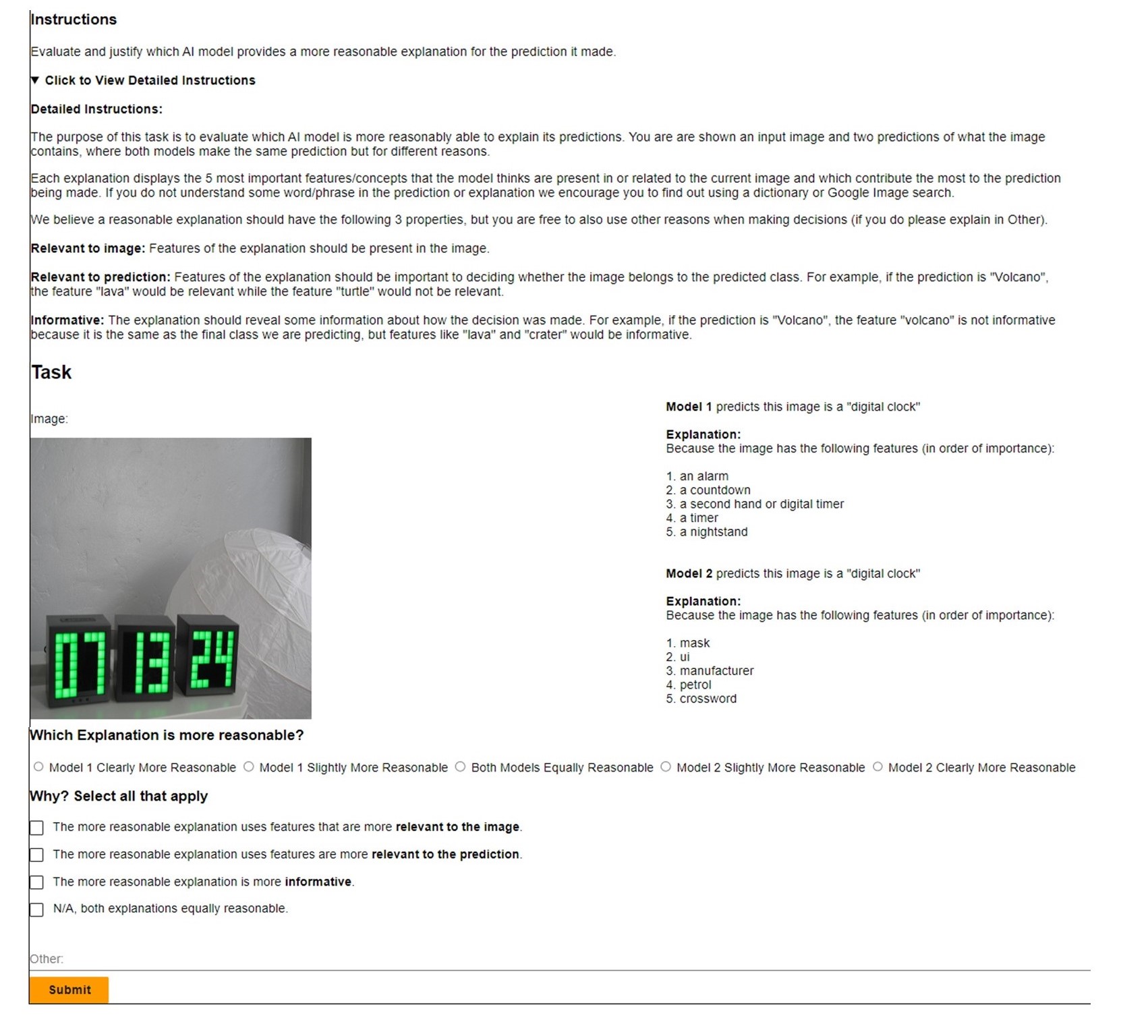}
    \caption{User Interface shown to Mechanical Turk users for Task 2. In this example Model 1 is LF-CBM while Model 2 is standard dense model.}
    \label{fig:task2_interface}
\end{figure}

\textbf{Setup:} We show workers explanations based on which concepts contributed the most to the decision similar to Figure \ref{fig:contributions}. However, we simplify this explanation to make it more suitable for crowdsourcing. Instead of displaying barplots and contribution magnitudes, we only show the names of 5 most contributing concepts. In addition, to avoid confusion we only show concepts with positive activation and positive weights. In this task we evaluated pairs of explanations to see how LF-CBM explanations compare with explanations from baseline models. Users were asked to rate which explanation is more reasonable and why. The task interface is shown in Figure \ref{fig:task2_interface}.  We compared our ImageNet LF-CBM against 3 baselines to ablate which parts of our model are most important for interpretability. The baselines are: 

\begin{itemize}
 \item Standard (dense) - ResNet50 trained on Imagenet
 \item Standard (sparse) - ResNet50 with sparse final layer
 \item LF-CBM (dense) - Our LF-CBM but with a dense final layer $W_F$
\end{itemize}

To create the explanations from standard models, we identified the neurons with highest contributions to the prediction, and used CLIP-Dissect~\citep{oikarinen2022clip} to get their concept. LF-CBM was compared against each baseline on 350 random input images, with 3 raters for each comparison for a total of 3150 comparisons. We only used inputs where all 4 models made the correct prediction, so the only difference was the explanations themselves.

We again used workers based in the US, with greater than 98\% approval rating and 10,000 HITs completed. Each worker was paid \$0.08 per comparison.

\textbf{Results:} 

\begin{figure}
    \centering    \includegraphics[width=0.9\textwidth]{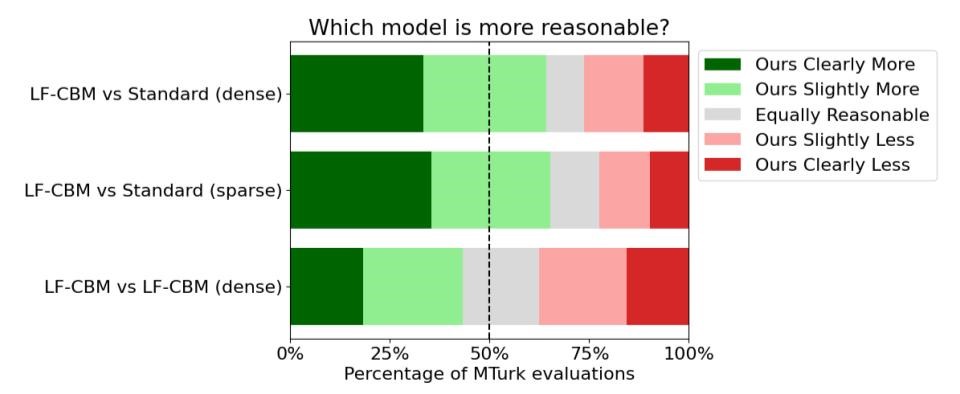}
    \caption{Crowdsourced results for Task 2, without filtering any responses. We can see users strongly prefer LF-CBM explanations over standard models (the first two rows), but sparsity has little effect (the 3rd row).}
    \label{fig:task2_res_all}
\end{figure}

\begin{figure}
    \centering
    \includegraphics[width=0.9\textwidth]{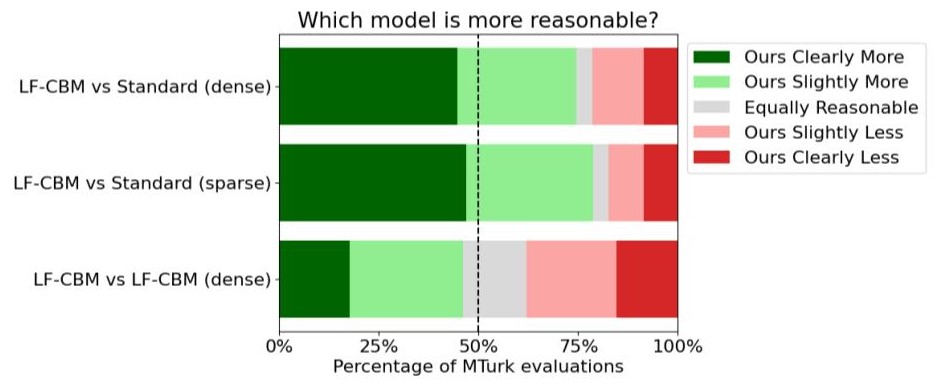}
    \caption{Crowdsourced results for Task 2, where we filter out responses from any worker who gave at least one inconsistent response. We can see user preference LF-CBM explanations over standard models even more clearly (the first two rows).}
    \label{fig:task2_res_filtered}
\end{figure}

The results of this experiment are displayed in Figures \ref{fig:task2_res_all} and \ref{fig:task2_res_filtered}. In Figure \ref{fig:task2_res_all} we can see users clearly preferred LF-CBM over Standard (dense) and Standard (sparse), with 64.2\% and 65.2\%(respectively) of users saying LF-CBM explanations are more reasonable (either slightly or clearly).

However, we found the responses were quite noisy for this task, and to improve their quality, we filtered away responses from any worker who gave an inconsistent response (for example selecting "Both Models Equally Reasonable" and "The More reasonable explanation is more informative" at the same time). The filtered results are shown in Figure \ref{fig:task2_res_filtered}. After filtering, the benefits of LF-CBM are even more clear, with 74.4\% of ratings saying LF-CBM explanations are more reasonable than Standard (dense) and 78.8\% are more reasonable than Standard (sparse). 

Interestingly, we found that sparsity did not really improve interpretability in this evaluation for either LF-CBM or the standard model. This was against our expectations, but we think it might be because this evaluation does not take into account how comprehensive the explanation is. A dense and sparse model often have similar top-5 most highly contributing concepts, but for dense concept the top-5 only explain a few percentage of the decision, while for sparse model they often explain 60-70\%, as seen in Figures \ref{fig:dense_cbm_explanation} and \ref{fig:imagenet_samples}.

As for why users found one or the other explanation more reasonable, the most common reason was "More Relevant to the Image"(selected ~80\% of the time), followed by "More Relevant to Prediction"(~60\%), while "More Informative" was the least popular, only selected around 30\% of the time.

\end{document}